\documentclass[twocolumn,secnumarabic,amssymb, nobibnotes, longbibliography, aps, prd]{revtex4-2}
\usepackage{amsmath, amsthm}
\usepackage{bm}
\usepackage{mathtools}
\usepackage{graphicx}
\usepackage{enumerate}
\usepackage{yhmath}
\usepackage[normalem]{ulem}
\usepackage{hyperref}
\usepackage{multirow}
\usepackage{caption}
\usepackage{subcaption}
\usepackage{xcolor}
\usepackage[usestackEOL]{stackengine}

\newcommand{\revtex}{REV\TeX\ }
\newcommand{\classoption}[1]{\texttt{#1}}
\newcommand{\macro}[1]{\texttt{\textbackslash#1}}
\newcommand{\m}[1]{\macro{#1}}
\newcommand{\env}[1]{\texttt{#1}}
\newcommand{\bfi}{\bfseries\itshape}

\newcommand{\mmathbb}[1]{{\mathbb{Z}_-}}
\newcommand{\ubar}[1]{\underline{#1}}
\newcommand{\uubar}[2]{\underline{#1}^{#2}}
\newcommand{\reg}{{\textup{reg}}}
\newcommand{\Volt}{\textup{Volt}}

\DeclareMathOperator*{\mathspan}{\textup{span}}
\DeclareMathOperator*{\argmin}{\arg\min} 

\newtheorem{theorem}{Theorem}

\newtheorem{lemma}[theorem]{Lemma}

\newtheorem{proposition}[theorem]{Proposition}

\newtheorem{example}[theorem]{Example}

\providecommand{\abs}[1]{\lvert#1\rvert}
\providecommand{\norm}[1]{\lVert#1\rVert}

\begin{document}
	
	\title{Infinite-dimensional next-generation reservoir computing}%
	
	\author{Lyudmila Grigoryeva}
	\email{Lyudmila.Grigoryeva@unisg.ch}
	\affiliation{Mathematics and Statistics Division, Universit\"{a}t Sankt Gallen, CH-9000 Switzerland}
	\author{Hannah Lim Jing Ting}
	\email{Hannahji001@e.ntu.edu.sg}
	\author{Juan-Pablo Ortega}
	\email{Juan-Pablo.Ortega@ntu.edu.sg}
	\affiliation{School of Physical and Mathematical Sciences, Nanyang Technological University, 637371 Singapore}
	
	\begin{abstract} 
		Next-generation reservoir computing (NG-RC) has attracted much attention due to its excellent performance in spatio-temporal forecasting of complex systems and its ease of implementation. This paper shows that NG-RC can be encoded as a kernel ridge regression that makes training efficient and feasible even when the space of chosen polynomial features is very large. Additionally, an extension to an infinite number of covariates is possible, which makes the methodology agnostic with respect to the lags into the past that are considered as explanatory factors, as well as with respect to the number of polynomial covariates, an important hyperparameter in traditional NG-RC. We show that this approach has solid theoretical backing and good behavior based on kernel universality properties previously established in the literature. Various numerical illustrations show that these generalizations of NG-RC outperform the traditional approach in several forecasting applications.
	\end{abstract}
	
	\maketitle
	
	\section{Introduction}
	\textit{Reservoir computing (RC)}~\cite{jaeger2001,maass1,Jaeger04,maass2}  has established itself as an important tool for learning and forecasting dynamical systems \cite{GHLO2014,Jaeger04,pathak:chaos,Pathak:PRL,Ott2018,wikner2021using,arcomano2022hybrid}. It is a methodology in which a recurrent neural network with a randomly generated state equation and a functionally simple readout layer (usually linear) is trained to proxy the data-generating process of a time series. One example is the \textit{echo state networks (ESNs)}~\cite{Matthews:thesis, Matthews1993, Jaeger04}, which have demonstrated excellent empirical performance and have been shown to have universal approximation properties in several contexts \cite{RC7, RC8, RC20, RC12}. Another family with similar theoretical properties is the {\it state-affine systems (SAS)} introduced in \cite{Sontag1979, RC6} and that is prevalent in engineering applications \cite{DangVanMien1984} and in quantum reservoir computing (QRC) \cite{QRC1, QRC2}.
	
	Despite the ease of training associated with the RC methodology, its implementation depends heavily on hyperparameters that are sometimes difficult to estimate robustly. This difficulty has motivated various authors to replace the standard RC approach with nonlinear vector autoregression in which the covariates are monomials constructed using the previous inputs. In these methods, the only hyperparameters to be chosen are the maximum order of the monomials and the number of lags of past signals. This approach has been called \textit{next-generation reservoir computing (NG-RC)}~\cite{gauthier2021next, bollt2021explaining, barbosa2022learning, kent2023controlling, NGRC24, NGRCPartialObs} and has displayed excellent performance in control and spatio-temporal forecasting tasks. 
	
	One downside of the NG-RC approach is that its performance and complexity depend strongly on the above-mentioned hyperparameters. Yet, as these hyperparameters grow, the computational effort associated with NG-RC increases exponentially. One goal of this paper is to place NG-RC in the context of kernel methods. Kernels are classical tools employed in static classification, regression, and pattern recognition tasks~\cite{schoelkopf:smola:book, Mohri:learning:2012, Steinwart2008}. Due to the Representer Theorem, kernels provide a way of passing inputs into higher-dimensional feature spaces where learning takes place linearly without scaling with the dimension of the feature space. By kernelizing NG-RC, we show a more computationally tractable approach to carrying out the NG-RC methodology that does not increase complexity with the hyperparameter values. In particular, we shall see that \textit{NG-RC is a particular case of polynomial kernel regression}.
	
	The idea that we just explained can be pushed all the way to considering all past lags and polynomial orders of arbitrarily high order in the polynomial kernel regression. Even though this leads to an infinite-dimensional covariate space (hence the title of the paper), the kernel regression can be explicitly and efficiently implemented using the recurrence properties of the {\it Volterra kernel} introduced in~\cite{RC25}. Since the Volterra kernel regression method has as covariates the left-infinite sequence of inputs and all the monomials of all degrees constructed from these inputs, this implies that, unlike NG-RC or the polynomial kernel regression, this approach is agnostic with respect to the number of lags and the order of monomials. Moreover, the Volterra kernel has been shown in~\cite{RC25} to be universal in the space of fading memory input/output systems with uniformly bounded inputs and, moreover, has a computational complexity that outperforms the NG-RC whenever higher-order monomials and lags are required for modeling more functionally complex systems. 
	
	The last part of the paper contains various numerical simulations that illustrate that {\bf (i)} using the more computationally tractable polynomial kernel regression, one has access to a broader feature space, which allows learning more complex systems than those typically presented when using the NG-RC methodology and {\bf (ii)} the Volterra kernel is a useful tool that can produce more accurate forecasts because it allows one to take infinite lags and monomial orders into account, which is of relevance in the modeling of long-memory phenomena that exhibit functionally complex dependencies. All the codes and data necessary to reproduce the results in the paper are available at~\cite{kernelngrcvolterra}.
	
	\section{Preliminary discussion}
	
	\subsection{Notation}
	Let $\mathbb{Z}$ denote the set of integers and $\mmathbb{Z}$ be the set of non-positive integers. Denote by $\mathbb{R}$ the set of real numbers and the $d$-dimensional Euclidean space by $\mathbb{R}^d$. We often work with sequence spaces. Let the space of sequences of $d$-dimensional real vectors indexed by $\mmathbb{Z}$ be denoted $\left(\mathbb{R}^d\right)^\mmathbb{Z}$. Given $\tau \in \mathbb{N}$, we shall use  $\left(\mathbb{R}^d\right)^\tau$ for the space of $\tau  $-long sequences of $d$-dimensional real inputs indexed by $\{-\tau+1, \ldots, -1, 0\}$. In the description of sequence spaces, we may also replace $\mathbb{R}^d$ with any subset of finite-dimensional Euclidean space $\mathcal{W}$. The components of sequences $\ubar{\mathbf{z}}\in \mathcal{W}^\mmathbb{Z}$ are given by ${\bf z} _i \in \mathcal{W}$, $i \in \mathbb{Z}_{-}$, that is, $\ubar{\mathbf{z}} = \left({\bf z} _i\right)_{i \in \mathcal{W}}$. Now, given $i \in \mathbb{Z}_{-}$, $\tau \in \mathbb{N}$, and $\mathbf{z}_i \in \mathcal{W}$, we denote by $\ubar{\mathbf{z}_i} \coloneqq (\ldots, \mathbf{z}_{i-1}, \mathbf{z}_{i}) \in {\cal W}^\mmathbb{Z}$ and by $\uubar{\mathbf{z}_i}{\tau} \coloneqq (\mathbf{z}_{i-\tau+1}, \ldots, \mathbf{z}_i) \in {\cal W}^\tau$. We use  $\mathcal{Z}$ and $\mathcal{Y}$ to refer to the input space and to the output space, respectively. The examples of  $\mathcal{Z}$ can be $\mathbb{R}^d$, $\left(\mathbb{R}^d\right)^\tau$, $\left(\mathbb{R}^d\right)^\mmathbb{Z}$, $\mathcal{W}^\mmathbb{Z}$, etc. The output space $\mathcal{Y}$ is typically finite-dimensional subsets of Euclidean space, or even subsets of $\mathbb{R}$.
	
	\subsection{Data generating processes}
	\label{sec: premise}
	Throughout this paper, we are interested in the learning and forecasting of discrete-time, time-invariant, and causal dynamic processes that are generated by functionals of the form $H : \left( \mathbb{R}^d \right)^\mmathbb{Z} \longrightarrow \mathcal{Y}$, that is, given $\ubar{{\bf z}}\in ({\Bbb R}^d)^\Bbb Z $, we have that ${\bf y}_t = H (\ubar{{\bf z}_t}) $, $t \in \Bbb Z $. In practice, the inputs $\ubar{{\bf z}}\in ({\Bbb R}^d)^\Bbb Z$ can be just deterministic sequences or realizations of a stochastic process. A related data generating process that we shall be using are the \textit{causal chains with infinite memory (CCIM)}  \cite{Dedecker2007a, doukhan2008weakly, alquier:wintenberger}. These are infinite sequences $(\mathbf{y}_i)_{i\in\mathbb{Z}}$, where $\mathbf{y}_i \in \mathcal{Y}$ for all $i \in \mathbb{Z}$, are such that $\mathbf{y}_i = H(\ubar{\mathbf{y}_{i-1}})$ for all $i\in\mathbb{Z}$ and some functional $H: {\cal Y} ^{\mathbb{Z}_{-}} \rightarrow {\cal Y} $. Takens' Theorem \cite{takensembedding} and its generalizations \cite{kocarev1995general, kocarev1996generalized, RC18, RC21} guarantee that the low-dimensional observations of dynamical systems follow, under certain conditions,  a dynamical prescription of this type.
	
	\subsection{Kernel methods}\label{sec: kernelmethods}
	
	{\bf Kernels and induced RKHS.} A \textit{kernel} on $\mathcal{Z}$ is a function $K : \mathcal{Z} \times \mathcal{Z} \longrightarrow \mathbb{R}$ that is symmetric and positive semi-definite. In this context, positive semi-definiteness means that for any $\alpha_i, \alpha_j \in \mathbb{R}$, $\mathbf{z}_i, \mathbf{z}_j \in \mathcal{Z}$, $i, j \in\{1, \ldots, n\}$, 
	\begin{equation*}
		\sum_{i=1}^n \sum_{j=1}^n \alpha_i \alpha_j K(\mathbf{z}_i, \mathbf{z}_j) \geq 0.
	\end{equation*}
	Given $\mathbf{z}_1, \ldots \mathbf{z}_n \in \mathcal{Z}$, define the \textit{Gram matrix} or {\it Gramian} to be the matrix $\mathsf{K} \coloneqq \left[K(\mathbf{z}_i, \mathbf{z}_j)\right]_{i, j=1}^n \in \mathbb{R}^{n \times n}$. The kernel function being symmetric and positive semi-definite is equivalent to the Gramian being positive semi-definite in a matrix sense. The next example is a family of kernel functions of importance in our discussion. 
	
	\begin{example}[Polynomial kernels]\label{eg: polykernel}
		\normalfont
		Let $\mathcal{Z} \subset \mathbb{R}^d$ for some $d \in \mathbb{N}$. For any constant $c>0$, a \textup{polynomial kernel} of degree $p\in\mathbb{N}$ is the function $K^{\textup{poly}} : \mathcal{Z} \times \mathcal{Z} \longrightarrow \mathbb{R}$ given by 
		\begin{equation*}
			K^{\textup{poly}}({\bf z}, {\bf z}') \coloneqq (\mathbf{z}^\top \mathbf{z}' + c)^p,\ \mathbf{z}, \mathbf{z}' \in \mathcal{Z}.
		\end{equation*}
	\end{example}
	
	Let $\mathbb{H}$ be a Hilbert space of real-valued functions on $\mathcal{Z}$ endowed with pointwise sum and scalar multiplication. Denote the inner product on $\mathbb{H}$ by $\langle \cdot, \cdot \rangle_\mathbb{H}$. We say that $\mathbb{H}$ is a \textit{reproducing kernel Hilbert space (RKHS)} associated to the kernel $K$ if the following two conditions hold. First, for all $\mathbf{z} \in \mathcal{Z}$, we have that the functions $K(\cdot, \mathbf{z}) \in \mathbb{H}$, and for all $\mathbf{z} \in \mathcal{Z}$ and all $f \in \mathbb{H}$, the \textit{reproducing property}, $f(\mathbf{z}) = \langle f, K(\cdot, \mathbf{z}) \rangle_\mathbb{H}$, $z \in {\cal Z} $, is satisfied. The maps of the form $K_\mathbf{z}(\cdot) \coloneqq K(\cdot, \mathbf{z}) : \mathcal{Z} \longrightarrow \mathbb{R}$ are called \textit{kernel sections}. Second, the Dirac functionals $\delta _{{\bf z}}: \mathbb{H} \rightarrow \mathbb{R} $ defined by $\delta _{{\bf z}}(f):=f({\bf z}) $ are continuous, for all ${\bf z} \in {\cal Z},$ with respect to the metric on $\mathbb{H}$ induced by the inner product $\langle \cdot, \cdot \rangle_{\mathbb{H}}$.
	
	\medskip 
	
	{\bf Canonical feature maps and spaces.} In this setup one may define the map $\Phi_{\mathbb{H}}: \mathcal{Z} \longrightarrow \mathbb{H}$ given by $\Phi_{\mathbb{H}}(\mathbf{z})\coloneqq K_\mathbf{z}$, ${\bf z}\in {\cal Z}$. Then from the reproducing property of the RKHS, the kernel function can be written as the inner product of kernel sections. Indeed, given ${\bf z}, {\bf z}' \in {\cal Z} $. we can write
	$$K(\mathbf{z}, \mathbf{z}') = \left<K_{\mathbf{z}'}, K_{\mathbf{z}} \right>_{\mathbb{H}}=\left< \Phi_{\mathbb{H}}(\mathbf{z}'), \Phi_{\mathbb{H}}(\mathbf{z}) \right>_{\mathbb{H}}.$$ 
	We call the map $\Phi_{\mathbb{H}}$ the \textit{canonical feature map} and the RKHS $\mathbb{H} $ is referred to as its {\it canonical feature space}.  
	
	By the Moore-Aronszajn Theorem~\cite{aronszajn1950theory, Steinwart2008}, any kernel has a unique RKHS, and this RKHS can be written as
	\begin{equation*}
		%\label{eq: rkhs_def}
		\mathbb{H} \coloneqq \overline{\mathspan \{ K_\mathbf{z} \mid \mathbf{z} \in \mathcal{Z} \}}, \\
	\end{equation*} 
	where the bar denotes the completion with respect to the metric induced by the inner product $\langle \cdot, \cdot \rangle_\mathbb{H}$. This inner product obviously satisfies that for $f = \sum^m_{i=1} \alpha_i K_{\mathbf{z}'_i}, g = \sum^n_{j=1} \beta_j K_{\mathbf{z}_j} \in \mathbb{H}$, $\langle f, g \rangle_{\mathbb{H}} = \sum_{i=1}^m \sum_{i, j=1}^n\alpha_i \beta_j K(\mathbf{z}'_i, \mathbf{z}_j)$. We use the symbol $\|\cdot\|_\mathbb{H}$ for the norm induced by $\langle \cdot, \cdot \rangle_\mathbb{H}$. 
	
	\medskip
	
	{\bf Feature maps and their associated kernels.} The construction we just presented produces an RKHS and a feature map out of a kernel map. Conversely, given any Hilbert space $(H, \left\langle \cdot , \cdot \right\rangle _H)$ and a map $\Phi: \mathcal{Z} \rightarrow H$ from the set ${\cal Z} $ into $H$, we can construct a kernel map that has $\Phi$ as a feature map and $H$ as a feature space, that is, define 
	\begin{equation}
		\label{representation with feature map}
		K(\mathbf{z}, \mathbf{z}'):=\left< \Phi(\mathbf{z}'), \Phi(\mathbf{z}) \right>_H, \quad {\bf z}, {\bf z}' \in \mathcal{Z}.
	\end{equation}
	Such a function $K$ is clearly symmetric and positive semi-definite; thus, it satisfies the conditions needed to be a kernel function. It can actually be proved (see~\cite[Theorem 4.16]{Steinwart2008}) that a map $K: \mathcal{Z} \times \mathcal{Z} \longrightarrow \mathbb{R}$ is a kernel if and only if there exists a feature map that allows $K$ to be represented as in \eqref{representation with feature map}. Note that given a kernel function, the feature representation \eqref{representation with feature map} is not unique in the sense that neither the feature space $H$ nor the feature map $\Phi: \mathcal{Z} \rightarrow H$ are unique. In particular, given $K$, the canonical feature map and the RKHS are a choice of feature map and feature space for $K$, respectively. 
	
	When a kernel function $K$ is defined as in \eqref{representation with feature map} using a feature map $\Phi: \mathcal{Z} \rightarrow H$, Theorem 4.21 in~\cite{Steinwart2008} proves that the corresponding RKHS $\mathbb{H}$ is given by
	\begin{equation}\label{eq:rkhs_feature}
		\mathbb{H} = \{f: \mathcal{Z} \rightarrow \mathbb{R} \mid 
		f(\cdot)=\left< w, \Phi(\cdot) \right>_H, w \in H \},
	\end{equation}
	equipped with the Hilbert norm 
	\begin{equation}
		\label{rkhs Hilbert norm}
		\left\|f\right\|_{\mathbb{H}}:=\inf \left\{\left\|w\right\|_{H}\mid w \in H \mbox{ such that } f(\cdot )= \left\langle w, \Phi(\cdot )\right\rangle_H\right\}.
	\end{equation}
	Using this fact, we prove the following lemma, which shows the equality of the RKHS spaces associated with kernels with feature maps related by a linear isomorphism. 
	\begin{lemma}\label{thm: rkhs_equal}
		Suppose $K_1$ and $K_2$ are kernels with feature spaces $H_1$ and $H_2$ and feature maps $\Phi_1 : \mathcal{Z} \rightarrow H_1$ and $\Phi_2 : \mathcal{Z} \rightarrow H_2$, respectively. Denote the corresponding RKHSs by $\mathbb{H}_1$ and $\mathbb{H}_2$. Suppose that there exists a bounded linear isomorphism $L: H_1 \rightarrow H_2$ such that $\Phi_2 = L \circ \Phi_1$, then $\mathbb{H}_1=\mathbb{H}_2$.
	\end{lemma}
	\begin{proof}
		By the bounded inverse theorem, $L^{-1}$, which is linear, is bounded. By the Riesz representation theorem, the adjoints of $L$ and $L^{-1}$, denoted by $L^*$ and $(L^{-1})^*$ are well defined. By~\eqref{eq:rkhs_feature}, for $f \in \mathbb{H}_1$, there exists $w \in H_1$ such that $f(\mathbf{z}) = \left< w, \Phi_1(\mathbf{z}) \right>_{H_1}$ for any $\mathbf{z} \in \mathcal{Z}$. Then it is easy to see that for any $\mathbf{z} \in \mathcal{Z}$, $f(\mathbf{z})=\left< (L^{-1})^* w, \Phi_2(\mathbf{z}) \right>_{H_2}$ so that $\mathbb{H}_1 \subset \mathbb{H}_2$. The reverse inclusion is similar. 
	\end{proof}
	
	\medskip
	
	{\bf Kernel ridge regression.} Suppose that ${\cal Z} \subset {\Bbb R}^d $, $\mathcal{Y} \subset \mathbb{R}$, and that a function $f : {\cal Z}\rightarrow {\cal Y} $ needs to be estimated using a finite sample of $n$ pairs of input and outputs $\{ (\mathbf{z}_i, y_i:=f(\mathbf{z}_i)) \}_{i=1}^n$ where $\mathbf{z}_i \in \mathcal{Z}$ and $y_i \in \mathcal{Y}$, $i=1, \ldots, n$. If the estimation is carried out using an \textit{empirical risk with squared loss} 
	\begin{equation*}
		\widehat{R}_n(g) := \frac{1}{n} \sum_{i=1}^{n} \left( g(\mathbf{z}_i) - y_i \right)^2,
	\end{equation*}
	and the hypothesis set is the RKHS $\mathbb{H} $ corresponding to a kernel $K$, the Representer Theorem~\cite{Mohri:learning:2012} states that the minimizer of the ridge regularized empirical risk on $\mathbb{H} $, that is,
	\begin{equation}\label{eq: representer_optproblem}
		f ^\ast =\argmin_{g\in\mathbb{H}} \left\{ \widehat{R}_n(g) + \Omega(\|g\|_\mathbb{H}^2) \right\},
	\end{equation}
	where $\Omega: (0, \infty) \longrightarrow \mathbb{R}$ is any increasing function, lies on the span of the kernel sections generated by the sample. More explicitly, the solution of \eqref{eq: representer_optproblem} has the form 
	\begin{equation}\label{eq: minimizer_representer}
		f^*(\cdot ) = \sum_{i=1}^n \alpha_i^\ast  K(\cdot, \mathbf{z}_i),
	\end{equation}
	where the vector $\bm{\alpha}^\ast  = (\alpha_1^\ast , \ldots, \alpha_n^\ast )^\top \in \mathbb{R}^n$ can be determined by solving the regularized linear (Gramian) regression problem (nonlinear in inputs, linear in covariates) associated to the Gram matrix $\mathsf{K}$ of the sample, that is,
	\begin{equation}\label{eq: kernel_optprob}
		\min_{\bm{\alpha} \in \mathbb{R} ^n} \left\{\frac{1}{n} \| \mathsf{K}\bm{\alpha} - \mathsf{Y}_n \|_2^2+ \Omega(\bm{\alpha} ^{\top} \mathsf{K} \bm{\alpha}) \right\},
	\end{equation}
	where $\mathsf{Y}_n:=(y _1, \ldots, y _n)^{\top}$ and $\|\cdot\|_2$ denotes the Euclidean norm. Let the regularization function $\Omega: (0, \infty) \longrightarrow \mathbb{R}$ be defined as $\Omega(u) = \lambda_{\textup{reg}} u$ for some regularization strength $\lambda_{\textup{reg}} > 0$. Then the optimization problem \eqref{eq: kernel_optprob} can be rewritten as
	\begin{equation}\label{eq: kernel_l2_optprob}
		\min_{\bm{\alpha} \in \mathbb{R} ^n} \left\{\frac{1}{n} \| \mathsf{K}\bm{\alpha} - \mathsf{Y}_n \|_2^2+ \lambda_{\reg} \bm{\alpha} ^{\top} \mathsf{K} \bm{\alpha} \right\},
	\end{equation} 
	which admits the closed form solution 
	\begin{equation}\label{eq: kernelclosedform}
		\bm{\alpha}^* = \left(\mathsf{K}^{2}+\lambda_\reg \mathsf{K} \right)^{-1} \mathsf{K} \mathsf{Y}_n.
	\end{equation}
	
	The optimization problem formulated in \eqref{eq: kernel_l2_optprob} is usually referred to as a {\it kernel ridge regression}. When the kernel $K$ that is used to implement it has a feature map $\Phi: {\cal Z}\rightarrow \left(H, \left\langle \cdot , \cdot \right\rangle_H\right)$ associated, the kernel ridge regression can be reformulated as a standard linear regression in which the covariates are the components of the feature map. More explicitly, due to \eqref{eq:rkhs_feature} and \eqref{rkhs Hilbert norm}, the following equality holds true:
	\begin{multline}
		\label{equality between optim}
		\min_{g\in\mathbb{H}} \left\{ \widehat{R}_n(g) + \lambda_\reg \|g\|_\mathbb{H}^2 \right\}\\=
		\min_{w\in {H}} \left\{ \widehat{R}_n(\left\langle w, \Phi(\cdot )\right\rangle_H) + \lambda_\reg \|w\|_{H}^2 \right\}.
	\end{multline}
	Then if we set $$w ^\ast =\argmin_{w\in {H}} \left\{ \widehat{R}_n(\left\langle w, \Phi(\cdot )\right\rangle_H) + \lambda_\reg \|w\|_{H}^2 \right\},$$ we can conclude that 
	\begin{equation}
		\label{kernelized vs nonlinear}
		f^* (\cdot )= \sum_{i=1}^n \alpha_i^\ast  K(\cdot, \mathbf{z}_i)= \left\langle w ^\ast , \Phi (\cdot )\right\rangle,
	\end{equation}
	which proves our claim in relation to interpreting the kernel ridge regression as a standard linear regression with covariates given by the feature map. In  particular,
	for new inputs $\mathbf{z}\in \mathcal{Z}$, the estimator $f^*$ generates the out-of-sample outputs 
	\begin{equation}\label{eq: kerneloutput}
		\widehat{y} = f^* (\mathbf{z}) = \sum_{i=1}^n \alpha^*_i K(\mathbf{z}, \mathbf{z}_i)=\left\langle w ^\ast , \Phi ({\bf z} )\right\rangle.
	\end{equation} 
	
	\medskip 
	
	\noindent {\bf Kernel universality.} An important feature of some kernels is their {\it universal} approximating properties \cite{micchelli2006universal, Steinwart2008, christmann2010universal} which we now define. Suppose we have a continuous kernel map $K$. For any compact subset $\mathcal{K} \subset \mathcal{Z}$, define the \textit{space of kernel sections} $K(\mathcal{K})$ to be the subset of $C^0(\mathcal{K})$ given by 
	\begin{equation}
		\label{def kernel sec compact}
		K(\mathcal{K}) \coloneqq \overline{ \mathspan \{ K_\mathbf{z} \mid \mathbf{z}  \in \mathcal{K} \} }.
	\end{equation}
	This time around, the bar denotes the uniform closure. Note that the continuity of $K$, the reproducing property, and the compactness of ${\mathcal K} $ imply that the uniform closure that defines $K \left({\mathcal K}\right) $ contains the completion of the vector space ${\rm span} \left\{K _{\mathbf{z}} \mid \mathbf{z} \in {\cal K}\right\}$. The continuous kernel $K$ is called \textit{kernel universal} when for any compact subset $\mathcal{K} \subset \mathcal{Z}$,
	\begin{equation*}
		K(\mathcal{K}) = C^0(\mathcal{K}).
	\end{equation*}
	
	Many kernels used in practice are universal. When a kernel is universal, the RKHS associated to the kernel, approximates arbitrarily well all continuous functions defined on compacta $\mathcal{K}$. This implies that having kernel universality ensures that the corresponding RKHS used in the kernel ridge regression problems discussed in the previous section is rich enough to approximate arbitrarily well any target (continuous) function. Examples of universal kernels include the Gaussian kernel and the Volterra kernel~\cite{RC25}, which we discuss in detail later on in Section~\ref{Infinite dimensional NG-RC and Volterra kernels}. On the other hand, the polynomial kernel, although popular, is not  universal~\cite{wang2013universalities}.
	
	Note that in the framework discussed in Section~\ref{sec: premise}, where the behavior of outputs is determined by causal functionals or CCIMs, whenever the corresponding functional is continuous and the inputs are defined on a compact input space, kernel universality implies that the elements of the corresponding RKHS can be used to uniformly approximate the data generating functional.
	
	\subsection{The NG-RC methodology}
	
	Next-generation reservoir computing, as introduced in \cite{gauthier2021next}, proposes to estimate a causal polynomial functional link between an explained variable $y _t$  at time $t\in \Bbb Z $ and the values of certain explanatory variables encoded in the components of  $\uubar{\mathbf{z}_t}{\tau} $ at time $t$ and in all the $\tau$-instants preceding it. Mathematically speaking, its estimation amounts to solving a linear regression problem that has as covariates polynomial functions of the explanatory variables in the past. 
	
	More explicitly, assume that we have a collection of $n$ inputs and outputs $\{ (\mathbf{z}_1, y_1), \ldots, (\mathbf{z}_n, y_n) \mid \mathbf{z}_i \in \mathbb{R}^d, y_i \in \mathbb{R} \}$. For each $t \in \{\tau, \ldots, n\}$, construct the $t$-th $\tau$-delay vector $\uubar{\mathbf{z}_t}{\tau}$, and the vector $\left[ \uubar{\mathbf{z}_t}{\tau} \right]^k$ containing all $k$-degree monomials of elements of $\uubar{\mathbf{z}_t}{\tau}$ as follows:
	\begin{align*} 
		\uubar{\mathbf{z}_t}{\tau} &\coloneqq 
		\left( 
		\begin{pmatrix}
			z_{t-\tau+1, 1} \\ \vdots \\ z_{t-\tau+1, d} 
		\end{pmatrix}, 
		\ldots, 
		\begin{pmatrix}
			z_{t, 1} \\ \vdots \\ z_{t, d} 
		\end{pmatrix} 
		\right) \in \left(\mathbb{R}^d\right)^\tau, \\
		\left[ \uubar{\mathbf{z}_t}{\tau} \right]^k 
		&\coloneqq 
		\left( \prod_{s=t}^{t-\tau+1} \prod_{u=1}^d z_{s, u}^{k_{s, u}} \right)_{\sum_{s, u} k_{s, u}=k,\hspace{0.1em} k_{s, u} \geq 0}.
	\end{align*}
	Re-indexing $\uubar{\mathbf{z}_t}{\tau}$ so that 
	\begin{equation}\label{eq: reindex_delay}
		\uubar{\mathbf{z}_t}{\tau} = 
		\left( 
		\begin{pmatrix}
			z_{t, 1} \\ \vdots \\ z_{t, d} 
		\end{pmatrix}, 
		\ldots, 
		\begin{pmatrix}
			z_{t, \tau d - d + 1} \\ \vdots \\ z_{t, \tau d} 
		\end{pmatrix}
		\right),
	\end{equation}
	we can then rewrite $\left[ \uubar{\mathbf{z}_t}{\tau} \right]^k$
	\begin{align*}
		\left[ \uubar{\mathbf{z}_t}{\tau} \right]^k =
		\left( \prod_{s=1}^{\tau d} z_{t, s}^{k_{t, s}} \right)_{\substack{\sum_{s=1}^{\tau d} k_{t, s}=k,\\ k_{t, 1}, \ldots k_{t, \tau d} \geq 0}},
	\end{align*}
	which we will use in the proof of Proposition~\ref{thm: ngrcispoly}.
	
	For a chosen maximum monomial order $p \in \mathbb{N}$, define the feature vector $\Phi: \left(\mathbb{R}^d\right)^{\tau} \longrightarrow \mathbb{R}^N$ as
	\begin{equation}
		\label{feature map poly}
		\Phi(\uubar{\mathbf{z}_i}{\tau}) = 
		(1, \left[\uubar{\mathbf{z}_i}{\tau}\right], \left[\uubar{\mathbf{z}_i}{\tau}\right]^2, \ldots, \left[\uubar{\mathbf{z}_i}{\tau}\right]^p)^\top,
	\end{equation}
	where $N$ denotes the dimension of the feature space, namely,
	\begin{equation}\label{eq: dim_ngrc}
		N = 1 + \sum_{k=1}^p \binom{\tau d + k - 1}{k} = \binom{\tau d + p}{p}.
	\end{equation}
	
	\medskip
	
	{\bf NG-RC ridge regression.} NG-RC proposes as link between inputs and outputs the solution of the ridge regression (nonlinear in inputs, linear in covariates) that uses the components of $\Phi$  as covariates, that is, it requires solving the optimization problem:
	\begin{equation} \label{eq: tik_problem}
		\argmin_{\mathbf{w} \in \mathbb{R}^N} \left\{ \frac{1}{n-\tau+1} \sum_{i=\tau}^n {\left(\Phi(\uubar{\mathbf{z}_i}{\tau})^\top \mathbf{w} - y_i \right)}^2 + \lambda_{\text{reg}} \| \mathbf{w} \|^2_2 \right\}.
	\end{equation} 
	
	This ridge-regularized problem obviously admits a unique solution that we now explicitly write. We first collect the feature vectors and outputs into a design matrix $\mathsf{X} $ and an output vector $ \mathsf{Y} $, respectively:
	\begin{equation*}
		\mathsf{X} \coloneqq 
		\begin{bmatrix}
			{\Phi(\uubar{\mathbf{z}_\tau}{\tau})}^\top \\ 
			\vdots \\ 
			{\Phi(\uubar{\mathbf{z}_n}{\tau})}^\top
		\end{bmatrix} \quad \text{ and } \quad
		\mathsf{Y} \coloneqq
		\begin{bmatrix}
			y_\tau \\ 
			\vdots \\ 
			y_n
		\end{bmatrix}.
	\end{equation*}
	The closed-form solution for the optimization problem \eqref{eq: tik_problem} is
	\begin{equation}\label{eq: ngrcsolution}
		\mathbf{w}^* = {(\mathsf{X}^\top \mathsf{X} + \lambda_\reg \mathbb{I}_{N})}^{-1} \mathsf{X}^\top \mathsf{Y}, 
	\end{equation}
	where $\mathbb{I}_N$ denotes the  identity matrix of dimension $N$.  Consequently, the output $y _t \in \mathbb{R}$ of an NG-RC system at each time step $t \in \Bbb Z$ corresponding to an input $\ubar{\mathbf{z}}\in \left(\mathbb{R}^d\right)^\Bbb Z$ is given by
	\begin{equation}\label{eq: ngrcoutput}
		\widehat{y}_t = {(\mathbf{w}^*)}^\top \Phi(\uubar{\mathbf{z}_t}{\tau}).
	\end{equation}
	
	\section{Kernelization of NG-RC}
	
	We now show that NG-RC can be kernelized using the polynomial kernel function introduced in Example \ref{eg: polykernel}. The term kernelization in this context means that, along the lines of what we saw in \eqref{kernelized vs nonlinear}, the solution \eqref{eq: ngrcoutput} of the NG-RC nonlinear regression problem can be written as the solution of the kernel ridge regression problem \eqref{eq: representer_optproblem} associated to the polynomial kernel.
	More explicitly, we shall see that the solution function  $f ^\ast (\uubar{\mathbf{z}}{\tau})={(\mathbf{w}^*)}^\top \Phi(\uubar{\mathbf{z}}{\tau}) $ in \eqref{eq: ngrcoutput} can be written as a linear combination of the kernel sections of $K^{\textup{poly}} $ generated by the data, with the coefficients obtained in \eqref{eq: kernel_l2_optprob} coming from the Representer Theorem. A similar result can be obtained for kernels based on Taylor polynomials as in~\cite{Steinwart2008}.
	
	\begin{proposition}\label{thm: ngrcispoly}
		Consider a sample of $n$ input/output observations $\{(\mathbf{z}_t, y_t)\}_{t\in\{1, \ldots, n\}}$ where $\mathbf{z}_t \in \mathbb{R}^d$ and $y_t\in\mathbb{R}$. Let $\tau \in \mathbb{N}$ be a chosen delay and let $p\in \mathbb{N}$ be a maximum polynomial order, let $K^{\textup{poly}} : \left(\mathbb{R}^d\right)^{\tau} \times \left(\mathbb{R}^d\right)^{\tau} \longrightarrow \mathbb{R}$
		\begin{equation}\label{eq: polykernel}
			K^{\textup{poly}}(\uubar{\mathbf{z}}{\tau}, \uubar{\mathbf{z}'}{\tau}) = (1 + (\uubar{\mathbf{z}}{\tau})^{\top} \uubar{\mathbf{z}'}{\tau})^p
		\end{equation}
		be the $\tau$-lagged polynomial kernel on $\mathbb{R}^{\tau d}$. Then,
		%\item [(i)] The map $\Phi: \left(\mathbb{R}^d\right)^{\tau} \longrightarrow \mathbb{R}^N$ introduced in \eqref{feature map poly} is a feature map for the kernel $K^{\textup{poly}} $, that is,
		%\begin{equation*}
		%K^{\textup{poly}}(\uubar{\mathbf{z}}{\tau}, \uubar{\mathbf{z}'}{\tau})= \Phi(\uubar{\mathbf{z}}{\tau})^{\top}\Phi(\uubar{\mathbf{z}'}{\tau}).
		%\end{equation*}
		the kernel regression problem on the left-hand side of \eqref{equality between optim}, corresponding to the input/output set $\{\left(\left(\uubar{\mathbf{z}_\tau}{\tau}\right), y_\tau\right), \ldots, \left(\left(\uubar{\mathbf{z}_n}{\tau}\right), y_n\right)\}$ and the kernel $K^{\textup{poly}}$ has the same solution as the NG-RC optimization problem given in \eqref{eq: tik_problem}, corresponding to $\{(\mathbf{z}_1, y_1), \ldots, (\mathbf{z}_n, y_n)\}$. In particular, the corresponding solution functions coincide, that is,
		\begin{equation}
			\label{solution functions}
			f ^\ast (\uubar{\mathbf{z}}{\tau})={(\mathbf{w}^*)}^\top \Phi(\uubar{\mathbf{z}}{\tau})=\sum _{i=0} ^{n- \tau}\alpha^\ast _i K^{\textup{poly}}_{\uubar{\mathbf{z}}{\tau}_{\tau+i}}(\uubar{\mathbf{z}}{\tau}), 
		\end{equation}
		for any $\uubar{\mathbf{z}}{\tau} \in  \left(\mathbb{R}^d\right)^{\tau}$ and where $\mathbf{w}^* \in \mathbb{R}^{N} $ is the NG-RC solution \eqref{eq: ngrcsolution}, $N$ as in \eqref{eq: dim_ngrc}, and $\boldsymbol{\alpha} ^\ast :=(\alpha _1^\ast , \ldots, \alpha^\ast _{n- \tau+ 1})$ is the solution of the Gramian regression in \eqref{eq: kernelclosedform} for $K^{\textup{poly}}$.
	\end{proposition}
	\begin{proof}
		For $i\in\{\tau,\ldots,n\}$, recall the reindexing of the $\tau$-delay vector $\uubar{\mathbf{z}_i}{\tau}$ in \eqref{eq: reindex_delay} and additionally let $z_{i, 0}=1$. By the multinomial theorem, for any $i, j \in \{\tau, \ldots, n\}$, 
		\begin{align}
			&K^{\textup{poly}}(\uubar{\mathbf{z}_i}{\tau}, \uubar{\mathbf{z}_j}{\tau}) 
			= \left(1 + (\uubar{\mathbf{z}_i}{\tau})^\top \uubar{\mathbf{z}_j}{\tau}\right)^p 
			= \left(\sum_{k=0}^{\tau d} z_{i, k} z_{j, k}\right)^p \notag\\
			&= \sum_{\substack{k_0+k_1,\ldots+k_{\tau d} = p \notag\\
					k_0, k_1, \ldots, k_{\tau d} \geq 0}} 
			\binom{p}{k_0, k_1, \ldots, k_{\tau d}} \prod_{t=0}^{\tau d} z_{i, t}^{k_t} z_{j, t}^{k_t} \notag\\
			&= 1 + \sum_{k_0=0}^{p-1} \sum_{\substack{k_1+\ldots+k_{\tau d}=p-k_0 \notag\\ k_1, \ldots, k_{\tau d} \geq 0}} \binom{p}{k_0, k_1, \ldots, k_{\tau d}} \prod_{t=1}^{\tau d} z_{i, t}^{k_t} \prod_{t=1}^{\tau d} z_{j, t}^{k_t} \notag\\
			&= \left( c \odot \Phi(\uubar{\mathbf{z}_i}{\tau}) \right)^\top \left( c \odot \Phi(\uubar{\mathbf{z}_j}{\tau}) \right),\label{feature map for poly}
		\end{align}
		where $\Phi: \left(\mathbb{R}^d\right)^{\tau} \longrightarrow \mathbb{R}^N$ is the NG-RC feature map introduced in \eqref{feature map poly}, $\odot$ denotes component-wise (Hadamard) multiplication, and the constant vector $c$ is given by
		\begin{equation*}
			c = \left( \sqrt{\binom{p}{k_o, k_1, \ldots, k_{\tau d}}} \right)^\top_{\substack{k_0 + k_1 + \ldots + k_{\tau d}=p \\ k_0, k_1, \ldots, k_{\tau d} \geq 0}} \in \mathbb{R}^N.
		\end{equation*}
		Notice that the relation \eqref{feature map for poly} implies that the map $c\odot\Phi: \left(\mathbb{R}^d\right)^{\tau} \longrightarrow \mathbb{R}^N$ is a feature map for the kernel $K^{\textup{poly}}$.
		Additionally, whenever $N < \infty$, the component-wise product with the vector $c\in \mathbb{R}^N$ can be written as a bounded linear isomorphism which can be represented by the diagonal $N$ by $N$ matrix with the elements of $c$ on the diagonal.

		Define the kernel function $K^{\textup{NG-RC}}: \left(\mathbb{R}^d\right)^{\tau} \times \left(\mathbb{R}^d\right)^{\tau} \longrightarrow \mathbb{R}$ obtained out of the dot product of the NG-RC feature vector \eqref{feature map poly}, that is,  
		\begin{equation*}
			K^{\textup{NG-RC}}(\uubar{\mathbf{z}}{\tau}, \uubar{\mathbf{z}'}{\tau}) \coloneqq \Phi(\uubar{\mathbf{z}}{\tau})^\top \Phi(\uubar{\mathbf{z}'}{\tau}),
		\end{equation*}
		which is obviously a kernel function because the dot product is symmetric and positive semi-definite.
		
		By the Moore-Aronszajn Theorem, $K^{\textup{poly}}$ and $K^{\textup{NG-RC}}$ each have unique RKHSs associated $\mathbb{H}_{\textup{poly}}$ and $\mathbb{H}_{\textup{NGRC}}$, respectively. Since each of their feature maps $c \odot \Phi$ and $\Phi$, respectively, are related by a bounded linear isomorphism, by Lemma~\ref{thm: rkhs_equal} we have that,
		\begin{equation*}
			%\label{equal rkhs}
			\mathbb{H}_{\textup{poly}}=\mathbb{H}_{\textup{NG-RC}}.
		\end{equation*}
		This implies that the kernel regression problems \eqref{eq: representer_optproblem}  associated with $K^{\textup{poly}}$ and $K^{\textup{NG-RC}}$ are identical and hence have the same solution. 
		%Denote now the Gram matrices of $K^{\textup{poly}}$ and $K^{\textup{NG-RC}}$ by $\mathsf{K}^{\textup{poly}}$ and $\mathsf{K}^{\textup{NG-RC}}$, respectively. Then by the equality \eqref{equal rkhs}, we have, 
		%\begin{multline*}
		%    \min_{\bm{\alpha} \in \mathbb{R}^{n-\tau+1}} \left\{\frac{1}{n-\tau+1} \| \mathsf{K}^{\textup{poly}}\bm{\alpha} - \mathsf{Y}_{n-\tau+1} \|_2^2+  \lambda_{\reg} \bm{\alpha} ^{\top} \mathsf{K}^{\textup{poly}}\bm{\alpha} \right\} \\
		%=\min_{\bm{\alpha} \in \mathbb{R}^{n-\tau+1}} \left\{\frac{1}{n-\tau+1} \| \mathsf{K}^{\textup{NG-RC}}\bm{\alpha} - \mathsf{Y}_{n-\tau+1} \|_2^2+  \lambda_{\reg} \bm{\alpha} ^{\top} \mathsf{K}^{\textup{NG-RC}}\bm{\alpha} \right\}.
		%  \end{multline*}
	This observation, combined with the identity \eqref{equality between optim}, proves the statement. 
\end{proof}

\begin{example}\label{eg: ngrcispoly}
	\normalfont
	In the setup of the previous proposition, consider the case $d=1$, $\tau=2$, and $p=2$. In that situation, the two kernels in the previous discussion are defined as:
	\begin{multline*}
		K^{\textup{poly}} (\uubar{\mathbf{z}_i}{\tau}, \uubar{\mathbf{z}_j}{\tau})\\
		= (1+(\uubar{\mathbf{z}_i}{\tau})^\top \uubar{\mathbf{z}_j}{\tau})^2 
		= \left(1+(z_i, z_{i-1}) (z_j, z_{j-1})^\top \right)^2  \\
		= 1 + 2z_i z_j + 2 z_{i-1}z_{j-1} + z_i^2 z_j^2 + z_{i-1}^2 z_{j-1}^2 
		+ 2 z_i z_j z_{i-1} z_{j-1}
	\end{multline*}    
	and provided that the NG-RC map is given by:
	\begin{equation*}
		\Phi(\uubar{\mathbf{z}_i}{\tau})=(1, z_j, z_{j-1}, z_j^2, z_j z_{j-1}, z_{j-1}^2)^\top,
	\end{equation*}
	we have that
	\begin{multline*}
		K^{\textup{NG-RC}} (\uubar{\mathbf{z}_i}{\tau}, \uubar{\mathbf{z}_j}{\tau})=
		\Phi(\uubar{\mathbf{z}_i}{\tau})^\top \Phi(\uubar{\mathbf{z}_j}{\tau}) \\
		= 1 + z_iz_j + z_{i-1} z_{j-1} + z_i^2 z_j^2 + z_{i-1}z_{j-1}z_i z_j + z_{i-1}^2 z_{j-1}^2.
	\end{multline*}
	As we already pointed out in the proof of Proposition \ref{thm: ngrcispoly}, the same monomials appear in $K^{\textup{poly}} $ and $K^{\textup{NG-RC}} $, which only differ by constants. 
	
	Another important observation that is visible in these expressions is the difference in computation complexity between NG-RC and its kernelized version introduced in Proposition \ref{thm: ngrcispoly}. Recall that NG-RC produces the vector $\mathbf{w}^\ast  \in \mathbb{R}^N$ in \eqref{solution functions} while the polynomial kernel regression yields $\boldsymbol{\alpha}^\ast \in \mathbb{R}^{n- \tau+1} $. NG-RC is a nonlinear (on inputs) regression on the components of the (six-dimensional in this case) feature map $\Phi $, and to carry it out, all those components have to be evaluated at all the data points; on the contrary, the kernelized version only requires the evaluation of $K^{\textup{poly}} $ at the data points, which is computationally simpler. {\it The kernelized version of NG-RC is, hence, computationally more efficient.} This difference is even more visible as $\tau $, $d$, and $p$ increase since the dependence \eqref{eq: dim_ngrc} of the number of covariates in the NG-RC regression on those parameters makes them grow rapidly, while the behavior of the computational complexity of the polynomial kernel $K^{\textup{poly}} $ is much more favorable. This difference in computational performance between the two approaches will be more rigorously analyzed later in Section \ref{Time complexities}.
\end{example}

\section{Infinite-dimensional NG-RC and Volterra kernels}
\label{Infinite dimensional NG-RC and Volterra kernels}

Apart from the computational efficiency associated with the kernelized version of NG-RC, this approach allows for an extension of this methodology that would be impossible in its original feature map-based version. More explicitly, in this section, we will see that by pursuing the kernel approach, NG-RC can be extended to the limiting cases $\tau \rightarrow \infty $, $p \rightarrow \infty$, hence taking into account infinite lags into the past and infinite polynomial degrees in relation with the input series. This is a valuable feature in modeling situations in which one is obliged to remain agnostic with respect to $\tau$ and $p$. The natural tool to carry this out is the {\it Volterra kernel} introduced in~\cite{RC25}, which is, roughly speaking, an infinite lag and infinite monomial degree counterpart of the polynomial kernel and that we recall in the following paragraphs.

Let $\pi_t: \left(\mathbb{R}^d\right)^\mmathbb{Z} \longrightarrow \left(\mathbb{R}^d\right)$, $t\in \mathbb{Z}_+$,  such that $\pi_t(\ubar{\mathbf{z}_i}) = \mathbf{z}_{i-t}$ be the projection operator onto the $(-t)$-th term of a semi-infinite sequence. Given $\tau \in \mathbb{N} $, define the $\tau$-time-delay operator $T_\tau : \left(\mathbb{R}^d\right)^\mmathbb{Z} \longrightarrow \left(\mathbb{R}^d\right)^\mmathbb{Z}$ by $\pi_t \left(T_\tau(\ubar{\mathbf{z}})\right) \coloneqq \pi_{t-\tau}\left(\ubar{\mathbf{z}}\right)$ for all  $t\in\mmathbb{Z}$. Given $M>0$, define the space of $M$-bounded inputs by $K_M \coloneqq \{ \ubar{\mathbf{z}} \in (\mathbb{R}^d)^\mmathbb{Z} \mid \| \pi_t\left(\ubar{\mathbf{z}}\right) \|_2^2 \leq M, \text{ for all } t \in \mmathbb{Z} \}$. Choose $\theta > 0$ such that $\theta^2 M^2 < 1$ and choose some $\lambda$ such that $0<\lambda<\sqrt{1-\theta^2M^2}$. Define the \textit{Volterra kernel} $K^\Volt : K_M \times K_M \longrightarrow \mathbb{R}$ by the recursion
\begin{equation}\label{eq: volterrakernelrecursive}
	K^\Volt(\ubar{\mathbf{z}}, \ubar{\mathbf{z}'}) = 1 + \lambda^2  \frac{K^\Volt(T_1(\ubar{\mathbf{z}}), T_1(\ubar{\mathbf{z}'}))}{1-\theta^2\left< \pi_0(\ubar{\mathbf{z}}), \pi_0(\ubar{\mathbf{z}'}) \right>}.
\end{equation} 
The rationale behind this recursion is the definition of the Volterra kernel proposed in \cite{RC25} as the kernel associated with a feature map obtained as the unique solution of a certain state space equation in an infinite-dimensional tensor space. The recursion in that state space equation implies the defining recursion in \eqref{eq: volterrakernelrecursive}. Alternatively, the Volterra kernel can be introduced by writing the unique solution of \eqref{eq: volterrakernelrecursive}, namely:
\begin{widetext}
	\begin{equation}\label{eq: volterrakernel}
		K^\Volt(\ubar{\mathbf{z}}, \ubar{\mathbf{z}'}) = 1 + \sum_{\tau=1}^\infty \lambda^{2\tau} 
		\frac{1}{1-\theta^2 \langle \pi_0(\ubar{\mathbf{z}}), \pi_0(\ubar{\mathbf{z}'}) \rangle} 
		\frac{1}{1-\theta^2 \langle \pi_1(\ubar{\mathbf{z}}), \pi_1(\ubar{\mathbf{z}'}) \rangle} \cdots \frac{1}{1-\theta^2 \langle \pi_{\tau-1}(\ubar{\mathbf{z}}), \pi_{\tau-1}(\ubar{\mathbf{z}'}\rangle}.
	\end{equation}
\end{widetext}
It can be verified that the Volterra kernel is a kernel map on the space of semi-infinite sequences with real entries in the sense discussed in Section~\ref{sec: kernelmethods}. 

\medskip

\noindent {\bf The Volterra kernel as an infinite order and lag polynomial kernel.} Observe that the $\tau$-lagged polynomial kernel map \eqref{eq: polykernel} can be rewritten as
\begin{align*}
	&K^{\textup{poly}}(\uubar{\mathbf{z}_i}{\tau}, \uubar{\mathbf{z}_j}{\tau}) 
	= (1 + (\uubar{\mathbf{z}_i}{\tau})^\top \uubar{\mathbf{z}_j}{\tau})^p \\
	&= (1 + \mathbf{z}_i^\top\mathbf{z}_j + \ldots + \mathbf{z}_{i-\tau+1}^\top \mathbf{z}_{j-\tau+1})^p = \\
	&\sum_{k=0}^p 
	\underbrace{
		\sum_{\substack{k_0 + \ldots + k_{\tau-1} = k \\
				k_0, \ldots, k_{\tau-1} \geq 0}}
		\binom{p}{p-k, k_0, \ldots, k_{\tau-1}} \prod_{t=0}^{\tau-1} (\mathbf{z}_{i-t}^\top \mathbf{z}_{j-t})^{k_t}
	}_{(*)},
\end{align*}
where we notice that the term marked with $(*)$ is the sum of all monomials of order $k$ on the variables that appear in the inner products $\mathbf{z}_i^\top \mathbf{z}_j, \ldots, \mathbf{z}_{i-\tau+1}^\top \mathbf{z}_{j-\tau+1}$. This expression yields the polynomial kernel as a polynomial of some finite degree $p$ on the components of the input terms up to some finite lag $\tau$. 

Rewriting \eqref{eq: volterrakernel} using the geometric series, we have for $\ubar{\mathbf{z}_i}, \ubar{\mathbf{z}_j} \in K_M \subset (\mathbb{R}^d)^\mmathbb{Z}$, 
\begin{multline*}
	K^\Volt(\ubar{\mathbf{z}_i},\ubar{\mathbf{z}_j}) 
	= 1 + \sum_{\tau=1}^\infty \lambda^{2\tau} \prod_{t=0}^{\tau-1} \sum_{k=0}^\infty \left( \theta^2 \mathbf{z}_{i-t}^\top \mathbf{z}_{j-t} \right)^k \\
	= 1 + \sum_{\tau=1}^\infty \sum_{k=0}^\infty \lambda^{2 \tau} \theta^{2k}
	\underbrace{ 
		\sum_{\substack{k_0 + \ldots + k_{\tau - 1} = k \\
				k_0, \ldots, k_{\tau-1} \geq 0 }} \prod_{t=0}^{\tau - 1} (\mathbf{z}_{i-t}^\top \mathbf{z}_{j-t})^{k_t}
	}_{(*)},
\end{multline*}
where $(*)$ is again a sum of all monomials of order $k$ on variables similar to the expression for $K^{\textup{poly}}$. However, in contrast to the polynomial kernel, note that in this case, we are taking monomial combinations of arbitrarily high degree and lags with respect to $\ubar{\mathbf{z}_i}$ and $\ubar{\mathbf{z}_j}$. This implies that the Volterra kernel considers additional functional and temporal information about the input, which allows us to use it in situations where we have to remain agnostic about the number of lags and the degree of monomials that need to be used. 

\medskip

\noindent {\bf Infinite-dimensionality and universality.} The discussion above hints that the Volterra kernel can be understood as the kernel induced by the feature map \eqref{feature map for poly} associated with the polynomial kernel, but extended to an infinite-dimensional codomain capable of accommodating all powers and lags of the input variables. This statement has been made rigorous in~\cite{RC25}, where the Volterra kernel was constructed out of an infinite-dimensional tensor feature space, which, in particular, makes it universal in the space of continuous functions defined on uniformly bounded semi-infinite sequences. This implies that {\it any continuous data-generating functional with uniformly bounded inputs can be uniformly approximated by elements in the RKHS generated by the Volterra kernel.} This is detailed in the following theorem proved in ~\cite{RC25}. The statement uses the notation introduced in \eqref{def kernel sec compact}. 
\begin{theorem}\label{thm: volterrauniversality}
	Let $K^\Volt : K_M \times K_M \longrightarrow \mathbb{R}$ be the Volterra kernel given by \eqref{eq: volterrakernel} and let $K^\Volt(K_M)$ be the associated space of kernel sections. Then 
	\begin{equation*}
		K^\Volt(K_M) = C^0(K_M). 
	\end{equation*}
\end{theorem}
In contrast, the polynomial kernel (equivalently NG-RC) is not universal (see \cite{wang2013universalities}). These arguments suggest that the Volterra kernel should outperform polynomial kernel regressions and the NG-RC in its ability to, for example, learn complex systems. This will indeed be illustrated in numerical simulations in Section \ref{sec: numerics}.  

\medskip

\noindent {\bf Computation of Volterra Gramians.} Even though the Volterra kernel is defined in the space of semi-infinite sequences, in applications, only finite samples of size $n$ of the form $\{(\mathbf{z}_i, y_i)\}_{i\in\{1, \ldots, n\}}$ are available. In that situation, it is customary to construct semi-infinite inputs of the form $\ubar{\mathbf{z}_i} = (\ldots, 0,0, \mathbf{z}_1, \ldots, \mathbf{z}_{i-1}, \mathbf{z}_i) \in (\mathbb{R}^d)^\mmathbb{Z}$, for each $i\in\{1, \ldots, n\}$, and we then define for that sample the Volterra Gram matrix  $\mathsf{K}^\Volt \in \mathbb{R}^{n \times n}$ as 
\begin{equation*}
	\mathsf{K}^\Volt_{i, j} = K^\Volt(\ubar{\mathbf{z}_i}, \ubar{\mathbf{z}_j}), \quad i, j \in \{1, \ldots, n\}.
\end{equation*}
Due to the recursive nature of the kernel map introduced in \eqref{eq: volterrakernelrecursive}, the entries of the Gram matrix can be computed also recursively by
\begin{equation}\label{eq: volterraGramrecursive}
	\mathsf{K}^{\Volt}_{i, j} = 1 + \frac{\lambda^2 \mathsf{K}^\Volt_{i-1, j-1}}{1-\theta^2\left< \mathbf{z}_i, \mathbf{z}_j \right>},
\end{equation} 
where $\mathsf{K}^\Volt_{0, 0} = \mathsf{K}^\Volt_{i, 0} = 1/(1-\theta^2)$ for all $i\in\{1, \ldots, n\}$. 

We recall now that, due to the Representer Theorem, the learning problem \eqref{eq: representer_optproblem} associated with the squared loss can be solved using the Gramian that we just constructed by computing \eqref{eq: kernel_l2_optprob}. Moreover, the solution $f ^\ast  $ has the form  $f ^\ast (\cdot )=\sum_{i=1}^n \alpha_i ^\ast  K^\Volt_{\ubar{\mathbf{z}_i}}(\cdot)$, with $\alpha _1^\ast , \ldots, \alpha_n^\ast  $ given by \eqref{eq: kernelclosedform}.

For a newly available set of inputs $\mathbf{z}_{n+1}, \ldots, \mathbf{z}_{n+h}$, the estimator can be used to forecast outputs $\widehat{y}_{n+1}, \ldots, \widehat{y}_{n+h}$. Extend the Gram matrix to a rectangular matrix $\mathsf{K}^\Volt \in \mathbb{R}^{n, \times n+h}$ by using the recursion
\begin{equation*}
	\mathsf{K}^{\Volt}_{i, n+j} = 1 + \frac{\lambda^2 \mathsf{K}^\Volt_{i-1, n+j-1}}{1-\theta^2\left< \mathbf{z}_i, \mathbf{z}_{n+j} \right>}
\end{equation*}  
that can be initialized by $\mathsf{K}^\Volt_{0, n+j}=1/(1-\theta^2)$ for all $j\in \{1, \ldots, h\}$. Then the forecasted outputs  are
\begin{equation*}
	\widehat{y}_{n+j} = \sum_{i=1}^n \alpha_i^\ast  K^\Volt_{\ubar{\mathbf{z}_i}}(\mathbf{z}_{n+j}) = \sum_{i=1}^n \alpha_i ^\ast \mathsf{K}_{i, n+j}^\Volt, \,  j\in \{1, \ldots, h\}.
\end{equation*}

\section{Numerics}\label{sec: numerics}

\subsection{Data generating processes and experimental setup}
Simulations were performed, using each of the three estimators discussed in the paper, on three dynamic processes: the Lorenz autonomous dynamical system, the Mackey-Glass delay differential equation, and the Baba-Engle-Kraft-Kroner (BEKK) input/output system.

For the Lorenz and Mackey-Glass dynamical systems, the task consisted of performing the usual path-continuation. During training, inputs are the spatial coordinates at time $t$ and outputs are the $(t+1)$-th spatial coordinate, and estimators are trained on a collection of $n$ input/outputs $\{(\mathbf{z}_t, \mathbf{z}_{t+1})\}_{t=1}^n$. To test their performance, the estimators are run autonomously. That is, after seeing initial input $\mathbf{z}_{n+1}$,the outputs $\widehat{\mathbf{z}}_{n+2}, \ldots, \widehat{\mathbf{z}}_{n+h}$, for some forecasting horizon $h$, are fed back into the estimator as inputs. These outputs $\widehat{\mathbf{z}}_{n+j}$ for $j=2, \ldots, h+1$ are compared against the reserved set of testing values $\mathbf{z}_{n+j}$ for $j=2, \ldots, h+1$, unseen by the estimators.

For the BEKK input/output system, the goal is to perform input/output forecasting. That is, during training, each estimator is given a set of inputs $\{\mathbf{z}_t\}_{t=1}^n$ and fitted against a set of outputs $\{\mathbf{y}_t\}_{t=1}^n$. Then, during testing, given a new set of unseen inputs $\{\mathbf{z}_t\}_{t=n+1}^{n+h}$, the outputs of the estimator $\{\widehat{\mathbf{y}}_t\}_{t=n+1}^{n+h}$ are compared against the actual outputs $\{\mathbf{y}_t\}_{t=n+1}^{n+h}$, unseen by the estimator.

The Lorenz system is a three-dimensional system of ordinary differential equations used to model atmospheric convection~\cite{lorenz1963deterministic}. The following Lorenz system
\begin{align*}
	\dot{x} = -10 (x-y) \space, \quad
	\dot{y} = 28 x-y-xz \space, \quad
	\dot{z} = -\frac{8}{3} z+xy 
\end{align*}
with the initial conditions
\begin{equation*}
	x_0=0, \quad y_0=1, \quad z_0=1.05
\end{equation*}
was chosen. A discrete-time dynamical system was derived using Runge-Kutta45 (RK45) numerical integration with time-step 0.005 to simulate a trajectory with 15001 points. The first 5000 points were reserved for training. The remaining points were reserved for testing. Although we consider all coordinates, one could potentially consider reconstructing the Lorenz dynamical system out of partial observations. In this case, the $\tau$-delay would be especially important due to the celebrated Takens' embedding theorem~\cite{takensembedding}. Regardless, even in the fully observable case used in this paper, the Lorenz dynamical system lies within the premise discussed in Section~\ref{sec: premise}.

The Mackey-Glass equation is a first-order nonlinear delay differential equation (DDE) describing physiological control systems given by~\cite{mackey-glass:paper}. We chose the following instance of the Mackey-Glass equation 
\begin{equation*}
	\dot{z} = \frac{0.2z(t-17)}{1+{z(t-17)}^{10}}-0.1z(t)
\end{equation*}
with the initial condition function being the constant function $z(t) = 1.2$. To numerically solve this DDE, the delay interval was discretized with time step of 0.02, then the usual RK45 procedure was performed on the discretized version of the system. The resulting dataset was flattened back into a one-dimensional dataset, and to reduce the size of the dataset, the dataset was further spliced to take every 50-th data point. The final dataset consisted of 7650 points, and the first 3000 points were reserved for training. The remaining points were reserved to compare against the path-continued outputs of each estimator. Due to the discretization process, the differential equation becomes a system of equations where each $z_t$ is a function of past observations, as is assumed by our premise in Section~\ref{sec: premise}. 

The BEKK model is an input/output parametric time series model that is used in financial econometrics in the forecasting of the conditional covariances of returns of stocks traded in the financial markets~\cite{engle:bekk}. We consider $d$ assets and the BEKK(1, 0, 1) model for their log-returns $\mathbf{r}_t$ and associated conditional covariances $\Sigma_t = \textup{Cov}(\mathbf{r}_t \mid \mathbf{r}_{t-1}, \mathbf{r}_{t-2}, \ldots)$ given by
\begin{align*}
	\mathbf{r}_t &= \Sigma_t^{1/2} \mathbf{z}_t, \quad \mathbf{z}_t \sim \textup{IIDN}(0_d, \mathbb{I}_d) \\
	\Sigma_t &= CC^\top + A \mathbf{r}_{t-1} \mathbf{r}_{t-1}^\top A^\top + B  \Sigma_{t-1} B^\top,
\end{align*}
where the input innovations $\mathbf{z}_t$ are Gaussian IID, and the output observations are the conditional covariances $\Sigma_t$. The diagonal BEKK specification is chosen where $C$ is an upper-triangular matrix, and $A$ and $B$ are diagonal matrices of dimension $d$. It is known that there exists a unique stationary and ergodic solution whenever $A_{ii} > 0, |B_{ii}| < 1$ for $i=1, \ldots, d$, which expresses $\Sigma_t$ as a highly nonlinear function of its past inputs~\cite{Boussama2011} (as per our premise in Section~\ref{sec: premise}). Since the covariance matrices are symmetric, we only need to learn the outputs $\mathbf{h}_t = \textup{vech}\Sigma_t) \in \mathbb{R}^q$, $q \coloneqq d(d+1)/2$, where the vech operator stacks the columns of a given square matrix from the principal diagonal downwards. An existing dataset from~\cite{RC25} was used with input dimensions of 15 and output dimensions of 120. The dataset consisted of 3760 input and output points, the first 3007 were reserved for training, and the remaining 753 were reserved for testing. Since the output points were very small, to minimize loss of accuracy due to computational truncation errors, the output values were scaled by 1000. The training output data was further normalized so that each dimension would have 0 mean and variance of 1. 

Since NG-RC methodology does not typically require normalization, the NG-RC datasets were not normalized. For the polynomial kernel, kernel values could become too large and result in truncation inaccuracies, so the training inputs were normalized to have a maximum of 1 and a minimum of 0. Due to the construction of the Volterra kernel, the input sequence space into the kernel for a finite sample is truncated with zeros. We thus demean the training input data. Moreover, to avoid incurring truncation errors for $\theta$ and $\lambda$ values, the maximum Euclidean norm of the inputs $M$ is set to 1, by scaling the training input values. Note that for all estimators, normalization is always performed based only on information from the training values, then the testing data is shifted and scaled based on what was used for the training data. This prevents leakage of information such as the mean, standard deviation, maximum, minimum, etc., to the testing datasets.

\subsection{Time complexities}
\label{Time complexities}

Following~\cite[page 280]{Mohri:learning:2012}, we compute the time complexities for the NG-RC, polynomial kernel regression and Volterra kernel regression. We assume for both kernel regressions, as per the procedure used in the numerical simulations, that the usual Euclidean dot product was used. 

In each forecasting scheme, training involves computing the closed-form solutions \eqref{eq: ngrcsolution} for the NG-RC and \eqref{eq: kernelclosedform} for the polynomial and Volterra kernel regressions. {For the NG-RC, to compute $\mathsf{X}^\top \mathsf{X}$ takes $O(n N^2)$ steps, recalling the definition of $N$ given in \eqref{eq: dim_ngrc}. To compute matrix inversion in \eqref{eq: ngrcsolution}, is $O(N^3)$. The remaining matrix multiplications have complexities dominated by $O(n N^2)$. Thus, the final complexity for training weights in NG-RC is $O(nN^2 + N^3)$. On the other hand, for polynomial or Volterra kernel regressions, one needs to compute the kernel map for each entry of the Gram matrix. For the polynomial kernel, in view of \eqref{eq: polykernel}, this is $O(\tau d)$, and for the Volterra kernel map, in view of \eqref{eq: volterraGramrecursive}, this is $O(d)$. Then, to compute the Gram matrix is $O(n^2 d \tau)$ and $O(n^2 d)$ for polynomial and Volterra kernel regression, respectively. Forecasting involves computing \eqref{eq: ngrcoutput} for the NG-RC and \eqref{eq: kerneloutput} for the polynomial and Volterra kernel regressions. For each time step, the complexity is $O(N)$ for the NG-RC, $O(n\tau d)$ for the polynomial kernel regression, and $O(nd)$ for the Volterra kernel regression.}

{The combinatorial $N$ term for the NG-RC can be bounded above by $(p+\tau d)^\kappa$ where $\kappa=\min(p, \tau d)$. Thus, in big-O notation, the $N$ term can be replaced by $(p+\tau d)^\kappa$. It can then be seen that when the sample size is small, and when $\tau$ and $p$ need not be large, the NG-RC will be faster than the polynomial and Volterra kernels. However, as $\tau$ and $p$ grow, as is needed to learn more complex dynamical systems, the complexity for NG-RC grows exponentially, and the polynomial and Volterra kernel regressions will outperform the NG-RC significantly.} For each of the forecasting schemes, the complexities associated with the training and generation of a single prediction are given in Table~\ref{tab: bigO}. 
\begin{table}
	\caption{Time complexities for all forecasting schemes discussed in this paper. Denote $\kappa = \min(p, \tau d)$.}
	\begin{ruledtabular}
		\begin{tabular}{lccl}
			& Training             & Prediction     \\ \hline
			NG-RC       & $O(n {(p+\tau d)^{2\kappa}} + {(p+\tau d)^{3\kappa}})$ & $O({(p+\tau d)^\kappa})$         \\ 
			Polynomial  & $O({n^2} \tau d + n^3)$ & $O(n \tau d)$  \\ 
			Volterra    & $O(n^2 d + n^3)$ & $O(nd)$        \\ 
		\end{tabular}
	\end{ruledtabular}
	\label{tab: bigO}
\end{table}

\subsection{Cross-validation}
For each estimator, hyperparameters have to be selected. The hyperparameters that were cross-validated and the chosen values are given in Table~\ref{tab: params}.
\begin{table*}
	\caption{Chosen parameters for each estimator and dataset}
	\begin{ruledtabular}
		\begin{tabular}{ccccc}
			System & Estimator & Washout & Hyperparameters \\ \hline
			\multirow{3}{*}{Lorenz} & NG-RC      & {3} & $(\tau=3, p=2, \lambda_\reg=1\times10^{-7})$ \\ 
			& Polynomial & 6 & $({\tau=6, p=2}, \lambda_\reg=1\times10^{-6})$  \\ 
			& Volterra   & 100 & $(\lambda\approx0.286, \theta=0.3, \lambda_\reg=1\times10^{-10})$ \\
			\hline              
			\multirow{3}{*}{Mackey-Glass} & NG-RC      & {4} & $(\tau=4, p=5, \lambda_\reg=1\times10^{-7})$ \\ 
			& Polynomial & {17} & (${\tau=17, p=4}, \lambda_\reg=1\times10^{-5})$\\ 
			& Volterra   & 100 & $(\lambda\approx0.859, \theta=0.3, \lambda_\reg=1\times10^{-9})$ \\   
			\hline 
			\multirow{3}{*}{BEKK} & NG-RC      & {1} & $({\tau=1}, p=2, \lambda_\reg=0.1)$ \\ 
			& Polynomial & {1} & $({\tau=1}, p=2, \lambda_\reg=0.1)$ \\ 
			& Volterra   & 100 & $(\lambda=0.72, \theta=0.6, \lambda_\reg=1\times10^{-3})$ \\
		\end{tabular}
	\end{ruledtabular}
	\label{tab: params}
\end{table*}
Note that the washout was not cross-validated for. For both NG-RC and polynomial kernel regression, washout is the number of delays taken. For the Volterra kernel, a longer washout is needed to wash the effect of truncating input samples with zeros. A washout of 100 was sufficient to generate meaningful results both for the full training set and when the training sets were restricted during cross-validation. 

For the path-continuation tasks (Lorenz and Mackey-Glass), to select hyperparameters, cross-validation was performed by splitting each training set into training-validation folds that overlapped. During validation, path-continuation was performed and compared with the validation set. That is, the outputs of each estimator were fed back as inputs, and these autonomously generated outputs were compared with the validation set. With overlapping datasets, a smaller training set would be sufficient to create multiple training folds starting from different initial points, such that in each training fold, the estimator has sufficient time to capture dynamics during training. Then during the validation phase for each fold, for a good estimator, there would be dynamical evolution in the outputs generated by the estimator. For example, the estimator did not just fit the average. This leads to meaningful validation set errors which improves the selection process for optimal hyperparameters.

For the BEKK input/output forecasting task, the usual time-series training-validation folds were used. That is, the training dataset was split into equally sized sets where the $i$-th training fold was the concatenation of the first $i$ sets and the $i$-th validation fold was $(i+1)$-set. For input/output forecasting, where estimator sees, during forecasting, a new set of inputs, this method of cross-validating turned out to be sufficient for estimators to capture the dynamics of input and output variables. Note that cross-validation training and testing were made to mimic as closely as possible the actual task to be performed on the full training set, so normalization in each fold was also performed as would have been done on the full training dataset. 

The range of parameters cross-validated were chosen so that the regularization was performed over the same set of values. As detailed in the previous section, time complexity for NG-RC grows exponentially for larger lag and degree hyperparameters. It was thus impractical to cross-validate over a large range of parameters as with each increase in number of lag or maximum degree of monomials, the computational time would grow exponentially. Thus, only a smaller range of parameters could be cross-validated over. The polynomial kernel was cross-validated over a larger space of the same delay and degree hyperparameters. When cross-validating for the Lorenz system, with fully observable coordinates, only one delay suffices to reconstruct the dynamical system. However, in practice, a higher number of lags may offer superior predictive performance. Hence we allowed for up to 10 lags to cross-validate over in the case of Lorenz. For the rest of the datasets, up to 101 lags were cross-validated over. Such a large range of hyperparameters was possible because by Proposition~\ref{thm: ngrcispoly} and Example~\ref{eg: ngrcispoly} the polynomial kernel regression uses the same covariates but is faster when more covariates are considered. Finally, mean squared error was chosen to be the metric over which the best hyperparameters were chosen.  

\subsection{Results}
{Pointwise and climate metrics were used to evaluate the performance of the estimators. Pointwise metrics are distance functions that evaluate the error committed by estimators from time step to time step. Climate metrics, see also \cite{wikner2024stabilizing}, are performance metrics that evaluate whether the statistical or physical properties are similar to the true system. We also consider the valid prediction time for each estimator in Lyapunov time for the two chaotic attractors (Lorenz and Mackey-Glass).}  

{Denoting the true value $\mathbf{y}$, the predicted value $\widehat{\mathbf{y}}$, and the testing set size $h$, the following pointwise error metrics were chosen: the normalized mean squared error (NMSE) given by
	\begin{equation*}
		\textup{NMSE}(\mathbf{y}, \widehat{\mathbf{y}}) =  \frac{1}{d} \sum_{u=1}^d \frac{\sum_{i=1}^h (y_{u, i} - \widehat{y}_{u, i})^2}{\sum_{i=1}^h (y_{u, i} - \bar{y}_{u})^2},%\frac{\sum_{i=1}^h \| \mathbf{y}_i - \widehat{\mathbf{y}}_i \|_2^2}{\sum_{i=1}^h \|\mathbf{y}_i\|^2_2},
	\end{equation*}
	where $\bar{y}_{u}$ denotes the average of the $u$-th dimension of the vector $\mathbf{y}$ over time steps $i=1, \ldots, h$,
	the mean absolute error (MAE) given by
	\begin{equation*}
		\textup{MAE}(\mathbf{y}, \widehat{\mathbf{y}}) = \frac{1}{h} \sum_{i=1}^h \| \mathbf{y}_i - \widehat{\mathbf{y}_i} \|_1
	\end{equation*}
	where $\|\cdot\|_1$ is the 1-norm, the median absolute error (MdAE) defined as
	\begin{equation*}
		\textup{MdAE}(\mathbf{y}, \widehat{\mathbf{y}}) = \frac{1}{d} \sum_{u=1}^d \textup{median}\left( \{ |y_{u, i} - \widehat{y}_{u, i}| \}_{i=1}^h \right),
	\end{equation*}
	and the mean absolute percentage error (MAPE)
	\begin{equation*}
		\textup{MAPE}(\mathbf{y}, \widehat{\mathbf{y}}) = \frac{1}{hd} \sum_{u=1}^d \sum_{i=1}^h \frac{|y_{u, i} - \widehat{y}_{u,i}|}{\max(\varepsilon, |y_i|)}
	\end{equation*}
	for some very small $\varepsilon>0$.
	
	Whether the estimator replicated the climate of the true time series was measured by considering the difference in the true and estimated power spectral density (PSD) and the difference in distributions using the Wasserstein-1 distance. 
	
	The PSD is the Fourier transform of the autocovariance function and represents the time series in its frequency domain~\cite{Brockwell2002}. The PSD of each time series was estimated by periodograms computed using Welch's method, provided by \texttt{scipy.signal.welch}, with the Hann window. The number of points per segment was chosen by visual inspection for a balance between frequency resolution and error variance. When the PSD tapers off to zero after some frequency $F$, the remaining frequencies are not considered in the final difference. Finally, the PSD error (PSDE) is computed by taking
	\begin{equation*}
		\textup{PSDE}(\textup{PSD}, \widehat{\textup{PSD}}) = \sum_{u=1}^d \sum_{f=1}^F \frac{| \textup{PSD}_{u, f}  - \widehat{\textup{PSD}}_{u, f} |}{\textup{PSD}_{u, f}},
	\end{equation*}
	where PSD is the periodogram of the actual data and $\widehat{\textup{PSD}}$ is the periodogram of the estimated data. The subscript $u, f$ denotes the $f$-th term in the $\textup{PSD}$ sequence in the $u$-th dimension.
	
	The Wasserstein-1 distance or earth mover distance, arising out of optimal transport~\cite{villani2009:optimal1}, is a measure of the distance between probability measures and is given by 
	\begin{equation}\label{eq: w1_distance}
		W_1(\mu, \widehat{\mu}) = \inf_{\pi \in \Gamma(\mu, \widehat{\mu})} \int \| \mathbf{y} - \widehat{\mathbf{y}} \|_2 d \pi(\mathbf{y}, \widehat{\mathbf{y}}),
	\end{equation}
	where $\Gamma(\mu, \widehat{\mu})$ is the set of probability distributions whose marginals are $\mu$ and $\widehat{\mu}$ on the first and second factors respectively. $\mu$ and $\widehat{\mu}$ are the joint distributions of $\mathbf{y}$ and $\widehat{\mathbf{y}}$ respectively.
	
	The Wasserstein-1 distance was computed to compare the distributions of the true and estimated systems in our paper. For one-dimensional systems, the Wasserstein-1 distance can be computed using \texttt{scipy.stats.wasserstein\char`_distance} which uses the equivalent Cramer-1 distance, that is,
	\begin{equation*}
		W_1(\mu, \widehat{\mu}) = \int_\mathbb{R} \left| \textup{CDF}(y) - \widehat{\textup{CDF}}(y) \right| dy,
	\end{equation*}
	where $\mu, \widehat{\mu}$ are the probability distributions of $y$ and $\widehat{y}$ respectively while $\textup{CDF}$, $\widehat{\textup{CDF}}$ denote the cumulative distributive functions. For time series in $d$-dimensions, using \texttt{scipy.stats.wasserstein\char`_distance\char`_nd}, corresponds to solving the linear programming problem in~\eqref{eq: w1_distance}. It is noted that computing the Wasserstein distance for the multidimensional case is significantly more computationally expensive than in the one-dimensional case. In the case of the Lorenz dynamical system, sampling had to be performed to make using \texttt{scipy.stats.wasserstein\char`_distance\char`_nd} tractable.
	
	{Finally, we note that for the dynamical systems Lorenz and Mackey-Glass, the Lyapunov time, defined to be the inverse of the top Lyapunov exponent, is a timescale for which a chaotic dynamical system is predictable. In deterministic path-continuing tasks, which were carried out for the Lorenz and Mackey-Glass dynamical systems, we measure the valid prediction time percent, $T_\textup{valid}$, which is the Lyapunov time taken for the predicted dynamics to differ from the true dynamics by 20\%. A similar metric was used in~\cite{wikner2024stabilizing}.}
	
	{The performance of the estimators was measured in the following manner for the dynamical systems. First, the valid prediction time was computed. Then, the ceiling of the best-performing valid prediction time is taken. The point-to-point error metrics NMSE, MAE, MdAE, and MAPE are measured only up to $\lceil T_\textup{valid} \rceil$. Beyond the characteristic predictable timescale given by the Lyapunov time, it is not meaningful to measure the step-to-step error as the trajectories diverge exponentially according to the Lyapunov exponent. To determine if the climate is well replicated over the full testing dataset, the climate metrics PSDE and $W_1$ are computed for the full testing dataset, with the Lorenz needing to be sampled to compute $W_1$. The errors are reported in Table~\ref{tab: errors}.}
	\begin{table*}
		\caption{Error values for each estimator and dataset. A 20\% deviation is allowed for $T_\textup{valid}$. NMSE, MAE, MdAE, and MAPE are computed up to $\lceil T_\textup{valid} \rceil$ for Lorenz and Mackey-Glass. PSDE and $W_1$ are computed for the full dataset (Lorenz was sampled). PSDE for BEKK is scaled by $10^{-4}$.}
		\begin{ruledtabular}
			\begin{tabular}{ccccccccccc}
				System & Estimator & $T_\textup{valid}$  & NMSE & MAE & MdAE & MAPE  & PSDE & $W_1$ \\ \hline
				\multirow{3}{*}{Lorenz}       
				& NG-RC     & 7.178   & 0.379  & 1.827 & 0.0897  & 1.598  & 7.606 & 2.114 \\
				& Polynomial & 9.126   & 0.188 & 1.155 & 0.0377  & 1.230  & {\bf 7.169} & {\bf 1.683}\\
				& Volterra  & {\bf 10.566} & {\bf 0.0934} & {\bf 0.428} & {\bf 0.00385} & {\bf 0.222} & 8.325 & 1.982 \\ 
				\hline\multirow{3}{*}{Mackey-Glass} 
				& NG-RC     & 0.3      & 0.980  & 0.188  & 0.174   & 0.235   & 28.815 & 0.155  \\ 
				& Polynomial & 7.035   & 0.0699 & 0.0274 & 0.00751 & 0.0329   & 6.831  & 0.00147 \\
				& Volterra  & {\bf 8.305} & {\bf 0.0333} & {\bf 0.0162} & {\bf 0.00202} & {\bf 0.0190}   & {\bf 5.059}  & {\bf 0.00138} \\ 
				\hline \multirow{3}{*}{BEKK}         
				& NG-RC & ---  & 0.875 & 0.0204 & 0.0166 & 0.644  & 1.565 & 0.332 \\ 
				& Polynomial & ---  & 0.877 & 0.0204 & 0.0166 & 0.642  & 1.140 & 0.337 \\
				& Volterra  & ---  & {\bf 0.619} & {\bf 0.0170} & {\bf 0.0139} & {\bf 0.634}  & {\bf 0.963} & {\bf 0.319} \\       
			\end{tabular}
		\end{ruledtabular}
		\label{tab: errors}
	\end{table*}
	
	{In more complex systems such as Mackey-Glass and BEKK, the Volterra reservoir and polynomial kernel regressions easily outperform the NG-RC because they have access to much richer feature spaces. In second-order systems such as the Lorenz system, even though pointwise errors perform better than in the polynomial and Volterra kernel regressions, the climate metrics indicate that the NG-RC better captured the climate of the true Lorenz dynamical system. It could be that a lower-order system offered by the NG-RC acts as a better proxy for lower-order true dynamical systems, which accounts for the difference in climate performance. This difference in climate replication performance, however, is only slight. Observe that in Figure~\ref{fig: psde}, all estimators capture the power spectral density of the original system well, even if the Volterra kernel is slightly outperformed by the polynomial kernel. For even more complex systems, both the kernel regression methods can capture the climate of the true dynamical system but the same cannot be said for NG-RC. A similar story holds when one considers the Wasserstein-1 distance. The distributions are in Figure~\ref{fig: dist}. Even though the Wasserstein-1 distance for Volterra performs the poorest, the difference in distribution performance is still small, and the bulk of the distribution is still replicated. On the other hand, for complex systems such as the BEKK, the polynomial kernel and the NG-RC fail to replicate the climate of the original system completely.}
	\begin{figure*}[t]
		\rotatebox{90}{\bfseries Lorenz \strut}\raisebox{-0.5\height}{
			\begin{subfigure}[b]{0.3\textwidth}
				\stackinset{c}{}{t}{-.15in}{\textbf{Volterra}}{
					\includegraphics[width=\textwidth]{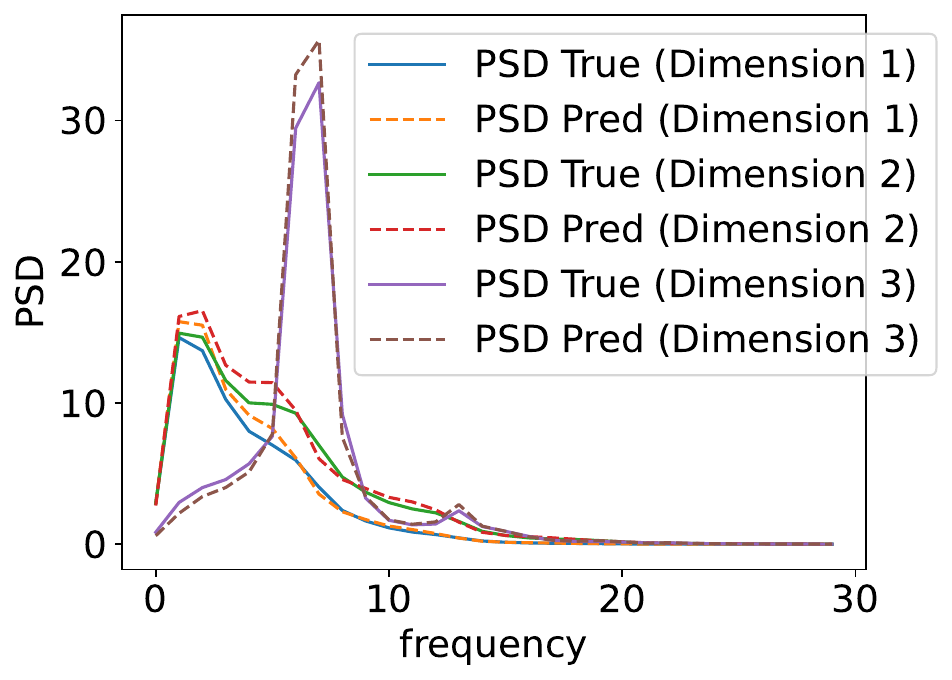}}
				\caption{}
				\label{fig: lorenz_volt_welch}
			\end{subfigure}
			\begin{subfigure}[b]{0.3\textwidth}
				\stackinset{c}{}{t}{-.15in}{\textbf{Polynomial}}{
					\includegraphics[width=\textwidth]{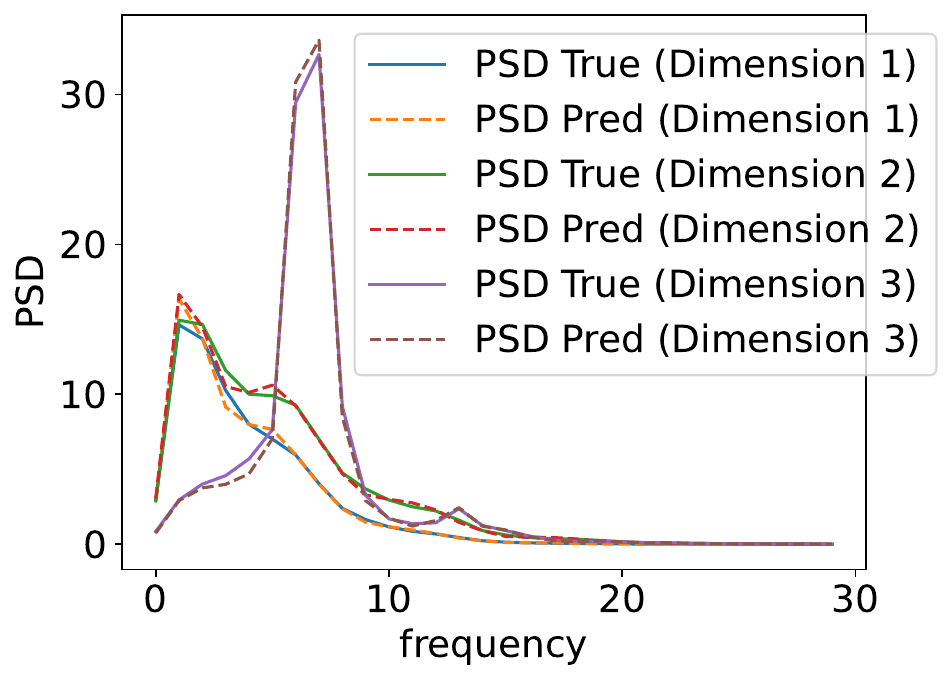}}
				\caption{}
				\label{fig: lorenz_poly_welch}
			\end{subfigure}
			\begin{subfigure}[b]{0.3\textwidth}
				\stackinset{c}{}{t}{-.15in}{\textbf{NG-RC}}{
					\includegraphics[width=\textwidth]{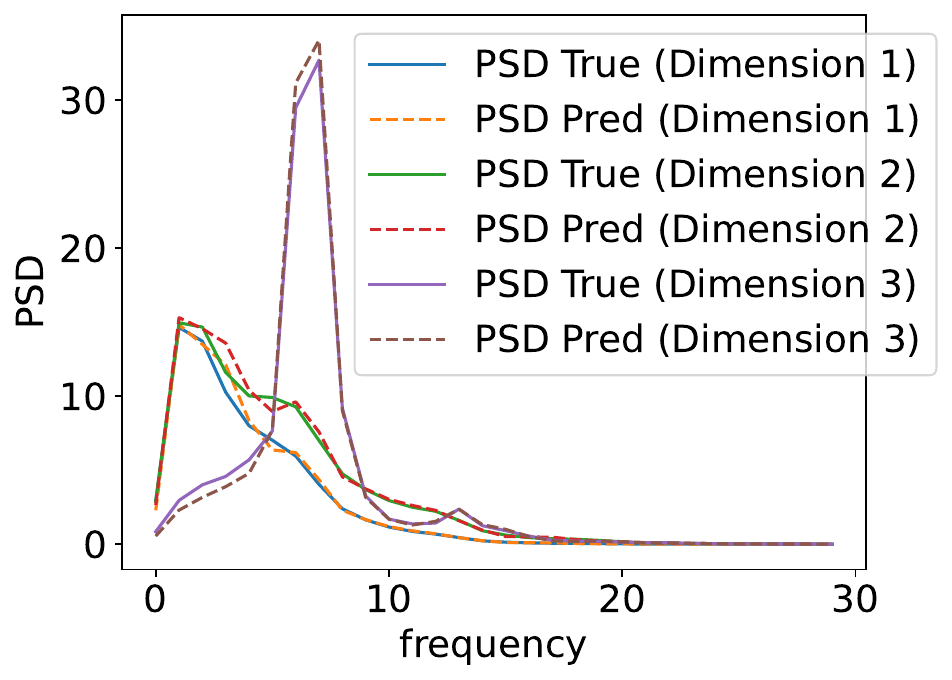}}
				\caption{}
				\label{fig: lorenz_ngrc_welch}
		\end{subfigure}}
		
		\rotatebox[origin=c]{90}{\bfseries Mackey-Glass \strut}
		\raisebox{-0.5\height}{
			\begin{subfigure}[b]{0.3\textwidth}
				\includegraphics[width=\textwidth]{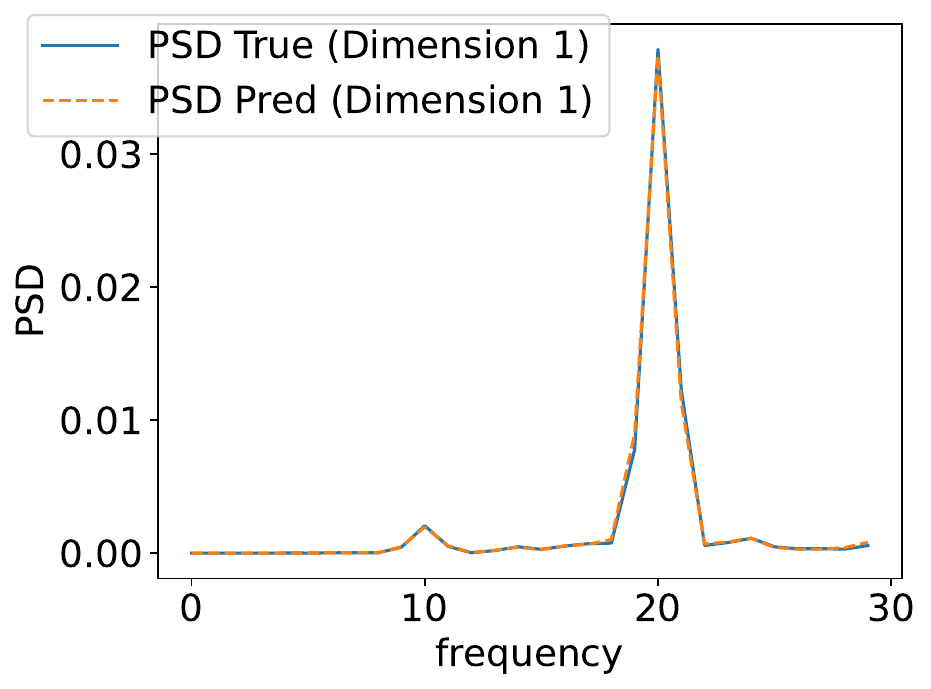}
				\caption{}
				\label{fig: mg_volt_welch}
			\end{subfigure}
			\begin{subfigure}[b]{0.3\textwidth}
				\includegraphics[width=\textwidth]{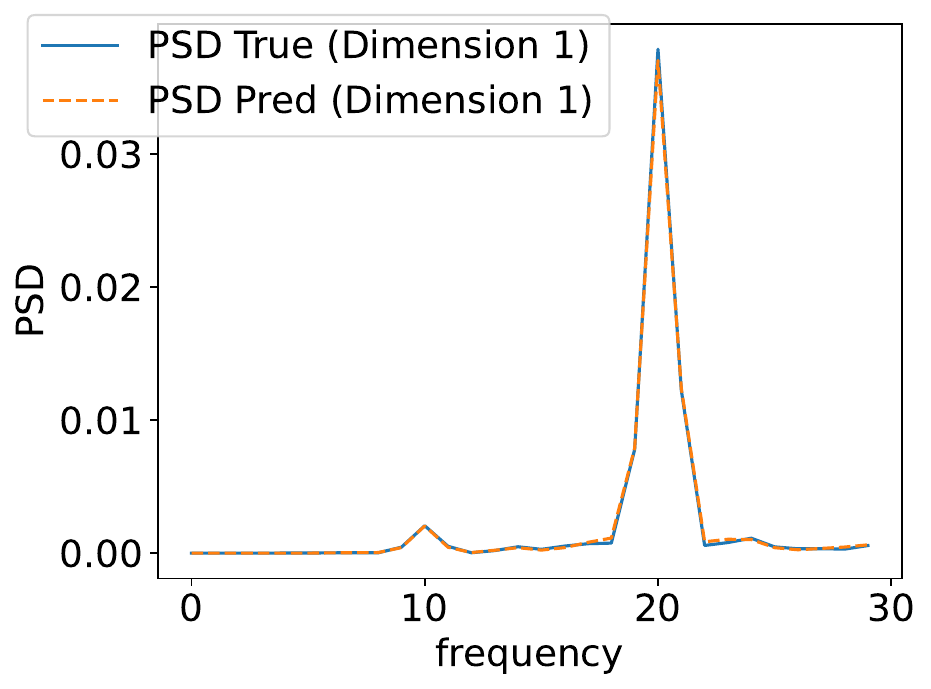}
				\caption{}
				\label{fig: mg_poly_welch}
			\end{subfigure}
			\begin{subfigure}[b]{0.3\textwidth}
				\includegraphics[width=\textwidth]{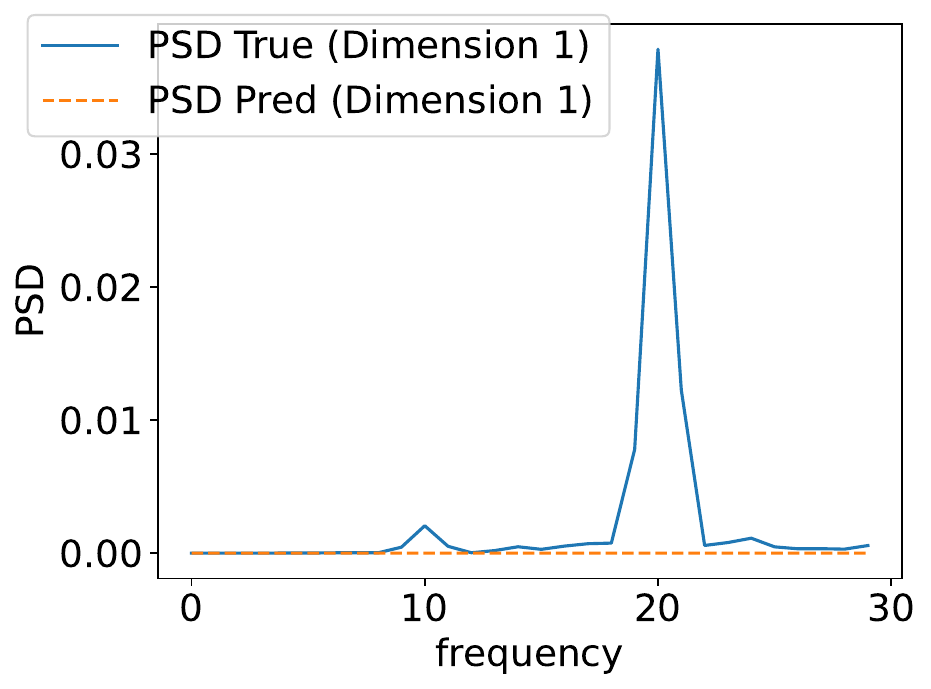}
				\caption{}
				\label{fig: mg_ngrc_welch}
		\end{subfigure}}
		
		\rotatebox[origin=c]{90}{\bfseries BEKK \strut}
		\raisebox{-0.5\height}{
			\begin{subfigure}[b]{0.3\textwidth}
				\includegraphics[width=\textwidth]{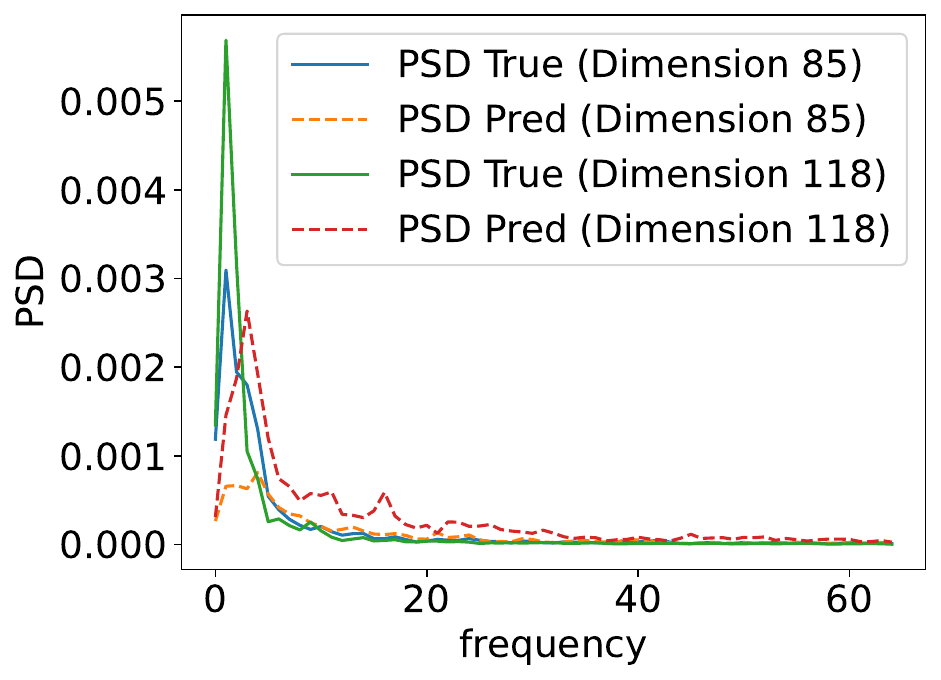}
				\caption{}
				\label{fig: bekk_volt_welch}
			\end{subfigure}
			\begin{subfigure}[b]{0.3\textwidth}
				\includegraphics[width=\textwidth]{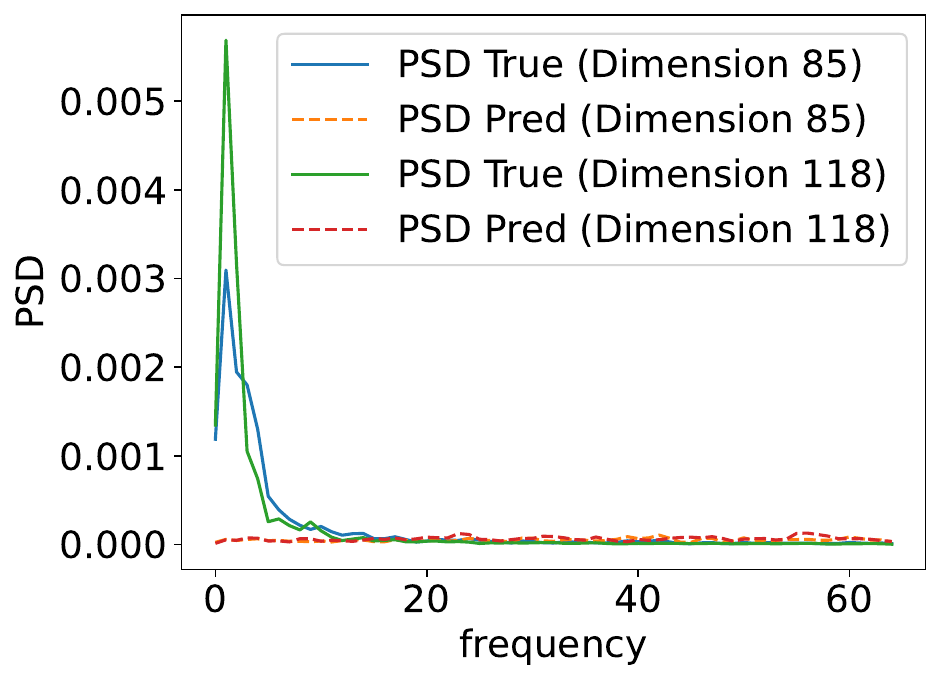}
				\caption{}
				\label{fig: bekk_poly_welch}
			\end{subfigure}
			\begin{subfigure}[b]{0.3\textwidth}
				\includegraphics[width=\textwidth]{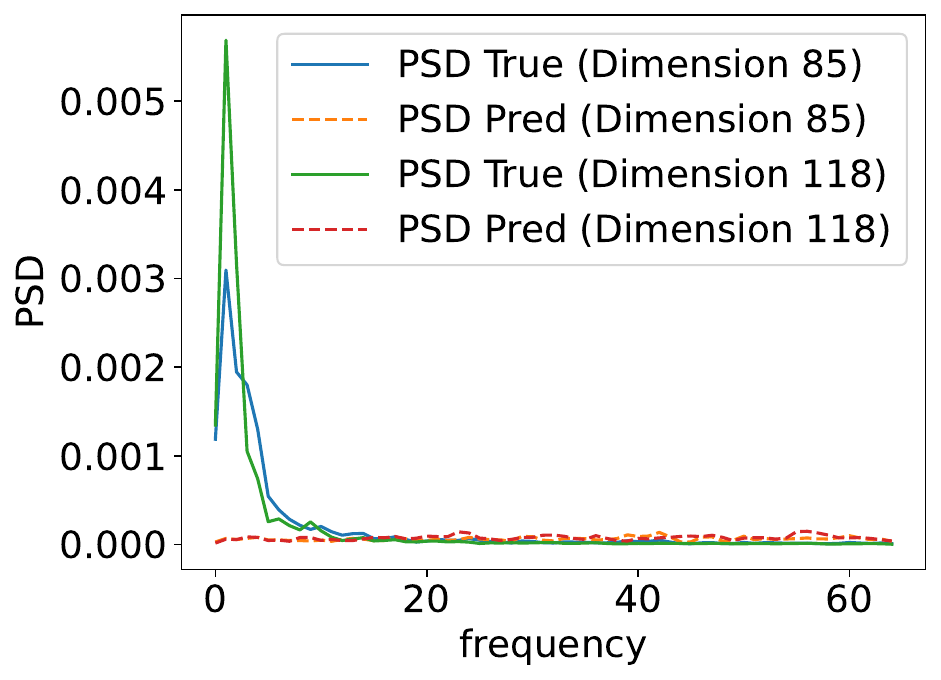}
				\caption{}
				\label{fig: bekk_ngrc_welch}
		\end{subfigure}}
		\caption{The PSD for: (\ref{fig: lorenz_volt_welch}) the Lorenz system for the Volterra kernel, (\ref{fig: lorenz_poly_welch}) the polynomial kernel, and (\ref{fig: lorenz_ngrc_welch}) NG-RC. Dimension $1$ corresponds to $x$, dimension $2$ to $y$, and dimension $3$ to $z$. PSD for Mackey-Glass system for Volterra kernel is (\ref{fig: mg_volt_welch}), polynomial kernel is (\ref{fig: mg_poly_welch}), and NG-RC is (\ref{fig: mg_ngrc_welch}). Dimension 1 refers to the $z$ value. PSD for BEKK for Volterra kernel is (\ref{fig: bekk_volt_welch}), polynomial kernel is (\ref{fig: bekk_poly_welch}), and NG-RC is (\ref{fig: bekk_ngrc_welch}). Only the two dimensions with the most visually prominent PSD are displayed (dimensions 85 and 118).}
		\label{fig: psde}
	\end{figure*}
	\begin{figure*}[t]
		\rotatebox{90}{\bfseries Lorenz \strut}\raisebox{-0.47\height}{
			\begin{subfigure}[b]{0.32\textwidth}
				\stackinset{c}{}{t}{-0.15in}{\textbf{Volterra}}{
					\includegraphics[width=\textwidth]{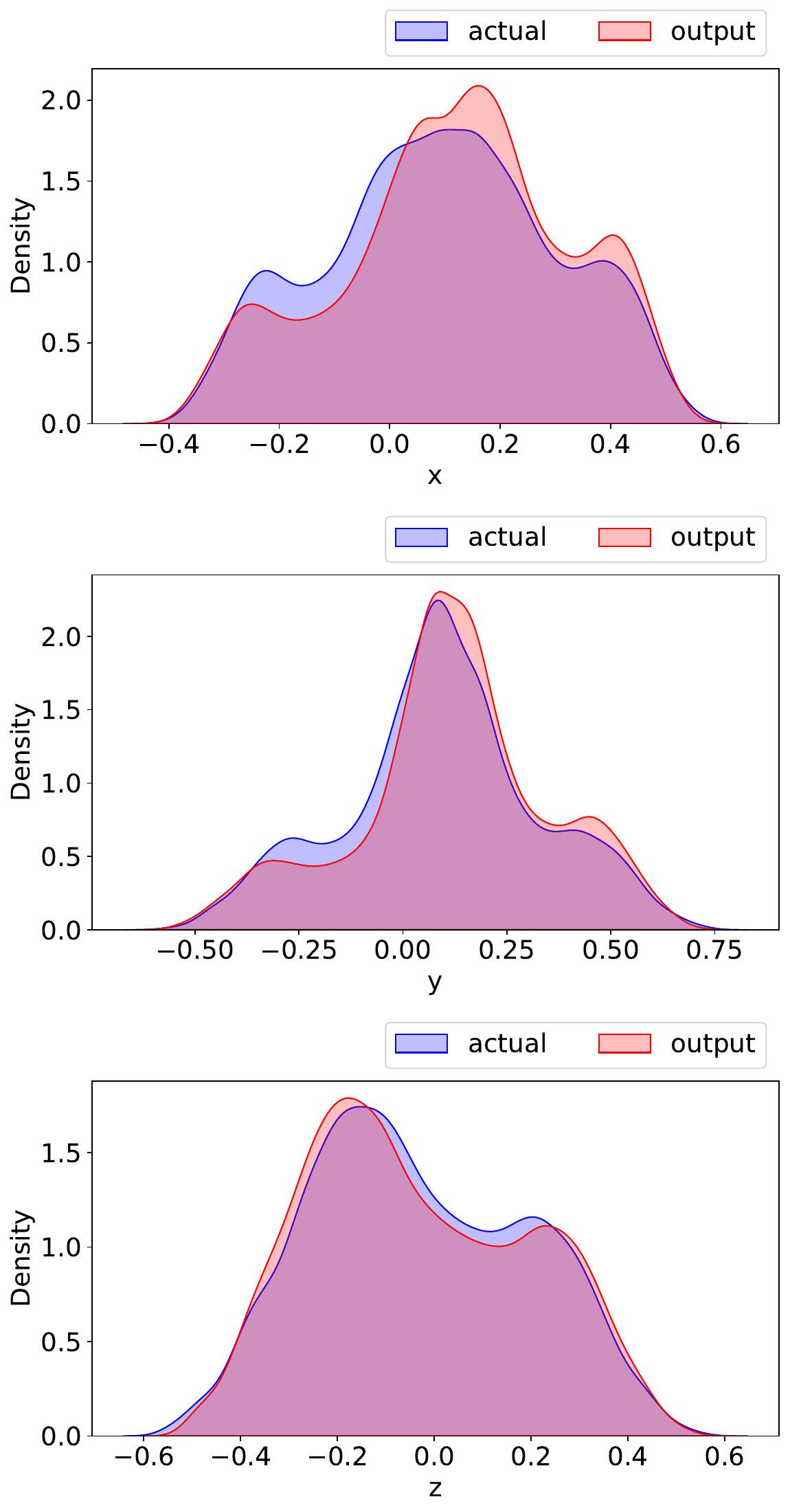}}
				\caption{}
				\label{fig: lorenz_volt_dist}
			\end{subfigure}
			\begin{subfigure}[b]{0.32\textwidth}
				\stackinset{c}{}{t}{-0.15in}{\textbf{Polynomial}}{
					\includegraphics[width=\textwidth]{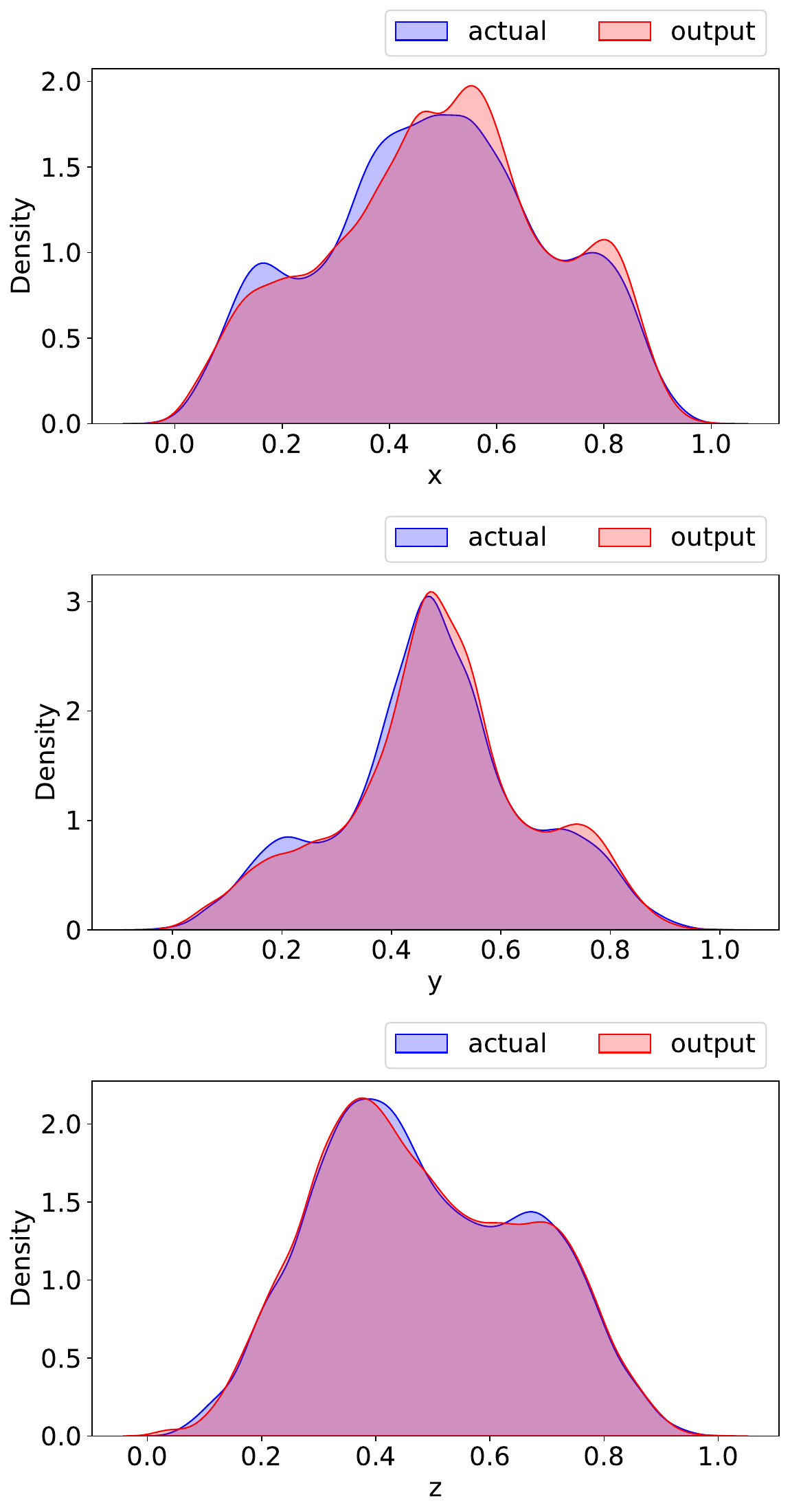}}
				\caption{}
				\label{fig: lorenz_poly_dist}
			\end{subfigure}
			\begin{subfigure}[b]{0.32\textwidth}
				\stackinset{c}{}{t}{-0.15in}{\textbf{NG-RC}}{
					\includegraphics[width=\textwidth]{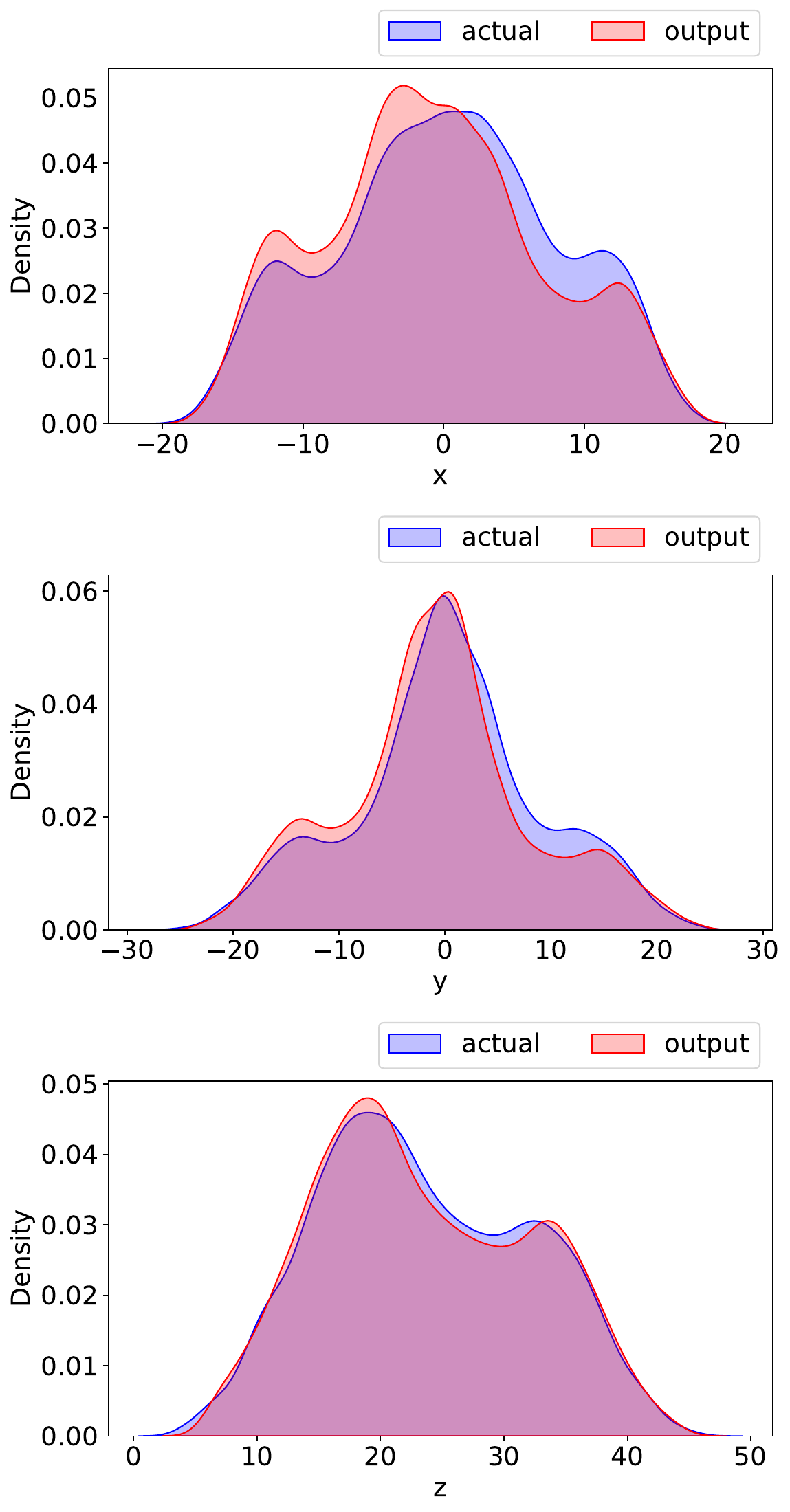}}
				\caption{}
				\label{fig: lorenz_ngrc_dist}
		\end{subfigure}}
		
		\rotatebox{90}{\bfseries Mackey-Glass \strut}\raisebox{-0.3\height}{
			\begin{subfigure}[b]{0.32\textwidth}
				\includegraphics[width=\textwidth]{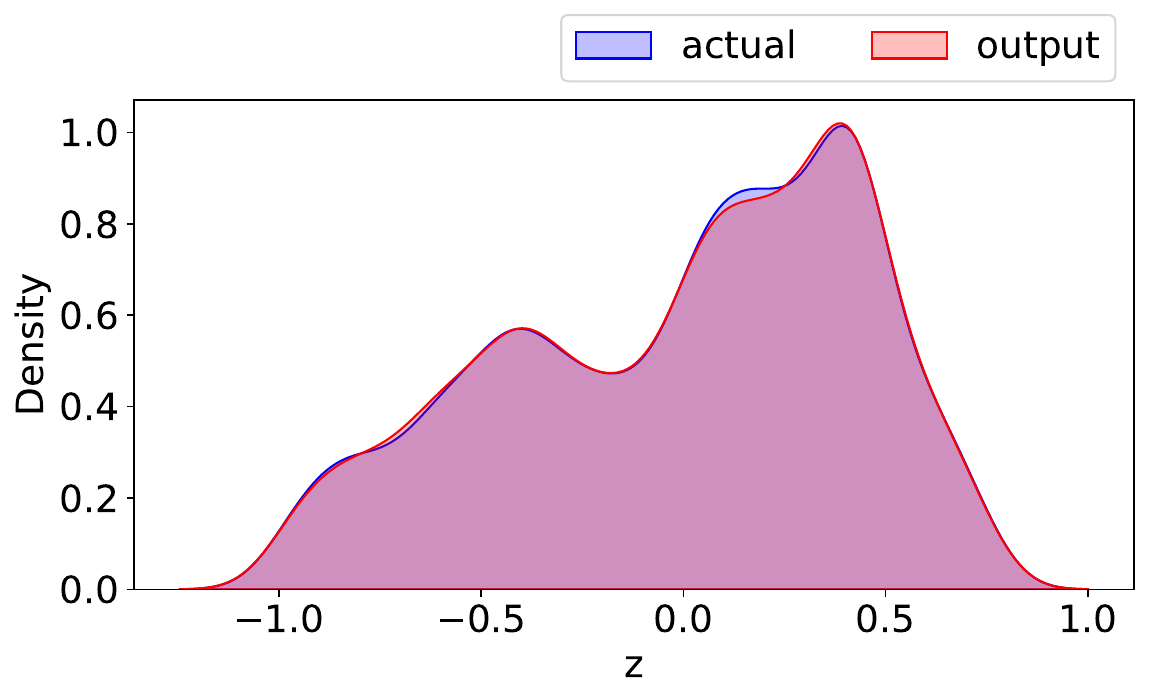}
				\caption{}
				\label{fig: mg_volt_dist}
			\end{subfigure}
			\begin{subfigure}[b]{0.32\textwidth}
				\includegraphics[width=\textwidth]{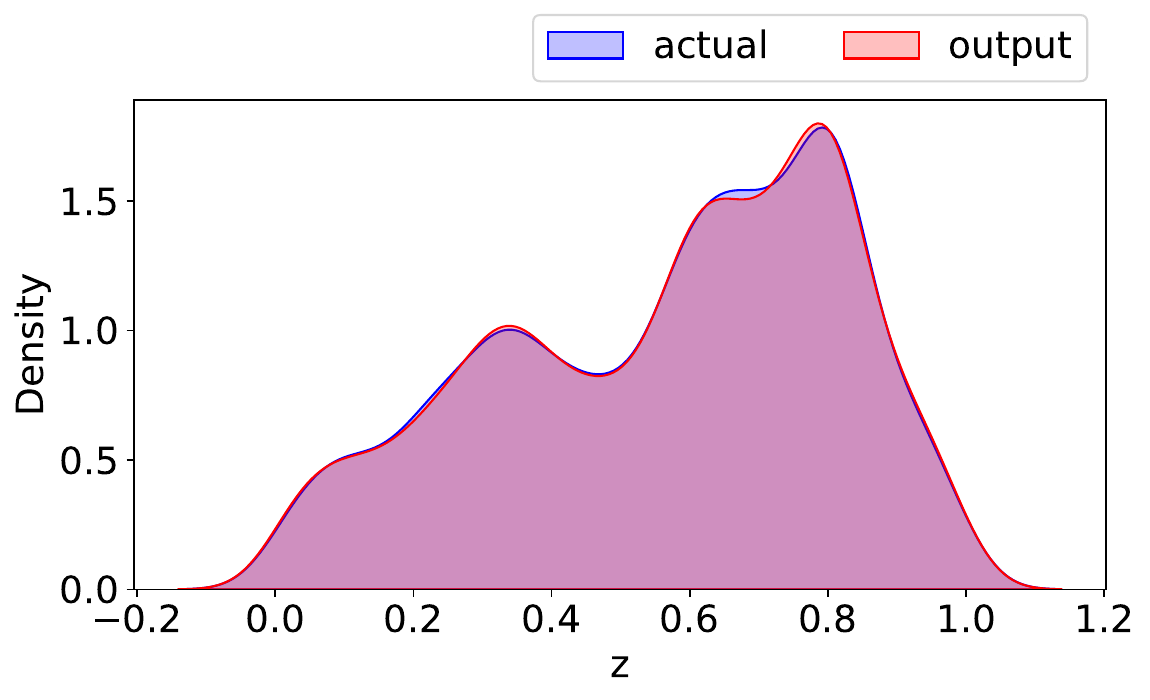}
				\caption{}
				\label{fig: mg_poly_dist}
			\end{subfigure}
			\begin{subfigure}[b]{0.32\textwidth}
				\includegraphics[width=\textwidth]{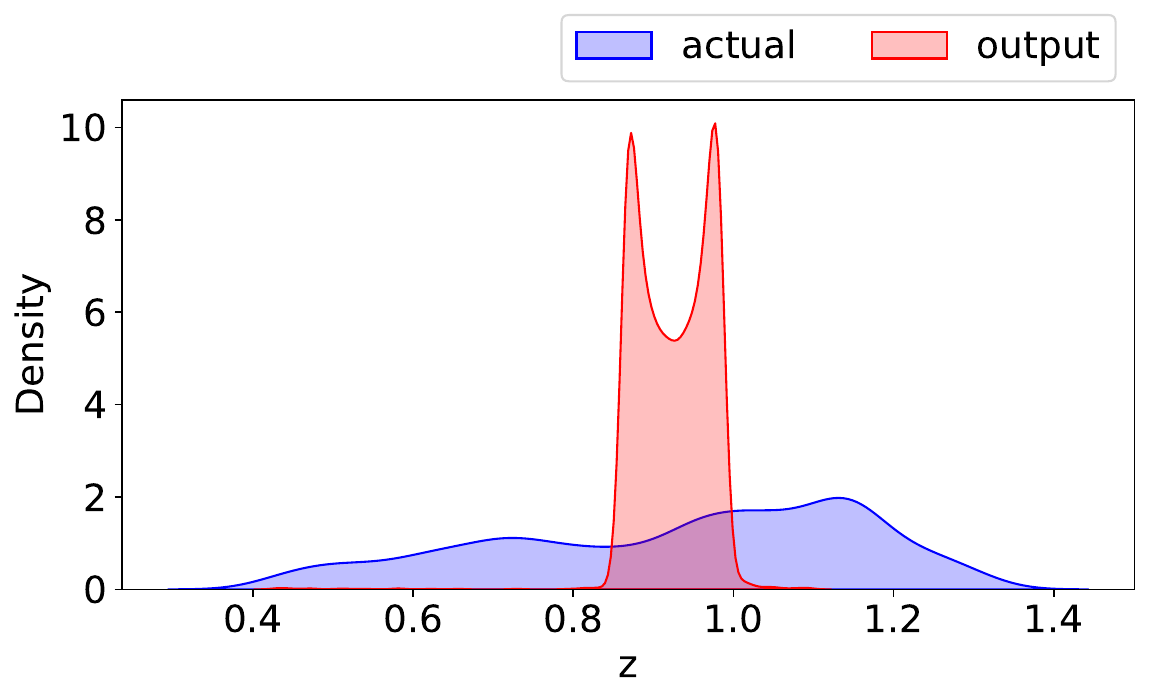}
				\caption{}
				\label{fig: mg_ngrc_dist}
		\end{subfigure}}
		
		\rotatebox{90}{\bfseries BEKK \strut}\raisebox{-0.45\height}{
			\begin{subfigure}[b]{0.32\textwidth}
				\includegraphics[width=\textwidth]{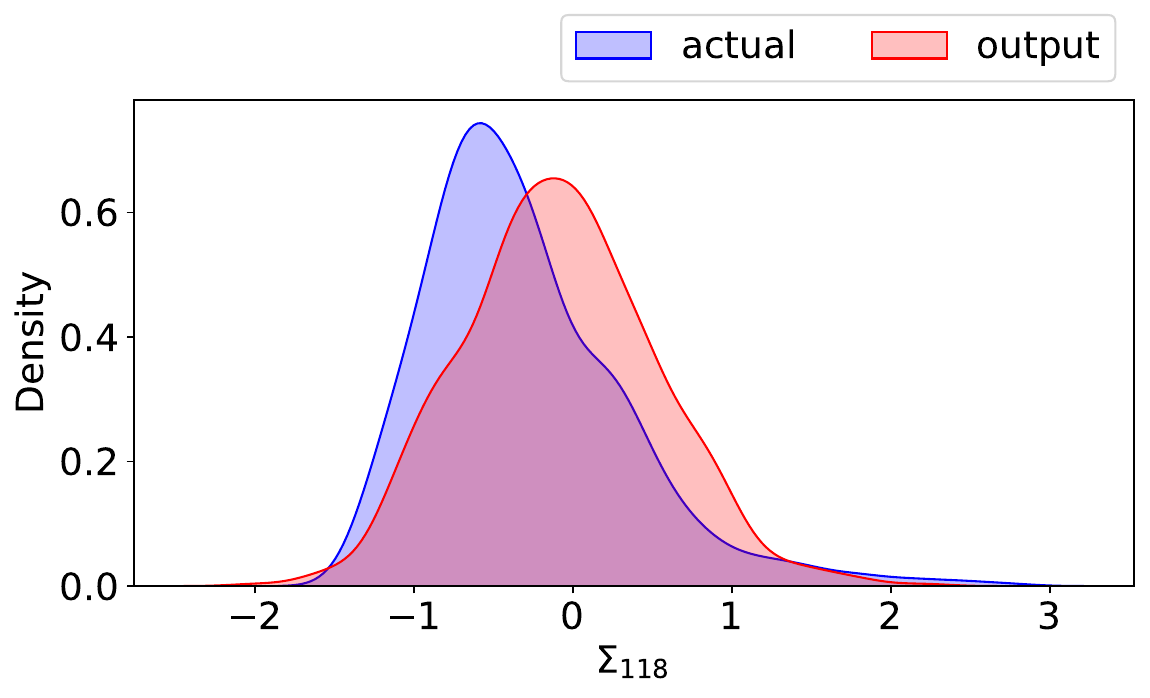}
				\caption{}
				\label{fig: bekk_volt_dist}
			\end{subfigure}
			\begin{subfigure}[b]{0.32\textwidth}
				\includegraphics[width=\textwidth]{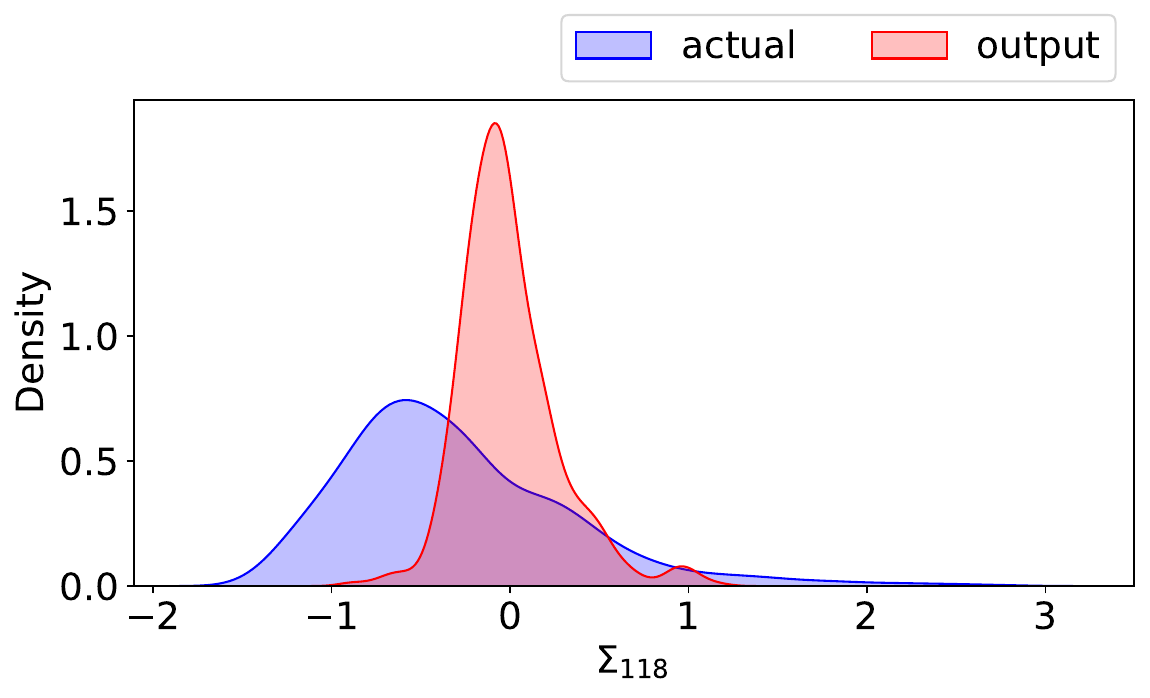}
				\caption{}
				\label{fig: bekk_poly_dist}
			\end{subfigure}
			\begin{subfigure}[b]{0.32\textwidth}
				\includegraphics[width=\textwidth]{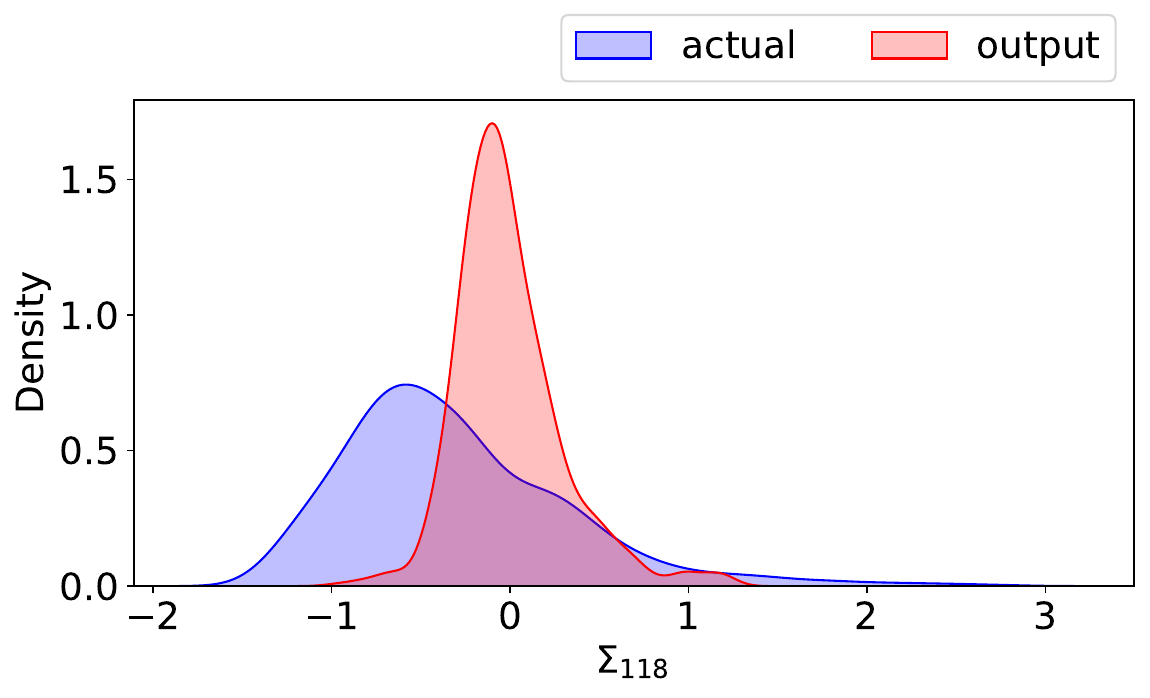}
				\caption{}
				\label{fig: bekk_ngrc_dist}
		\end{subfigure}}
		\caption{The distributions for Lorenz system for the Volterra kernel is in (\ref{fig: lorenz_volt_welch}), the polynomial kernel is in (\ref{fig: lorenz_poly_welch}), and NG-RC is in (\ref{fig: lorenz_ngrc_welch}). The distribution for the Mackey-Glass system for Volterra kernel is  (\ref{fig: mg_volt_welch}), the polynomial kernel is (\ref{fig: mg_poly_welch}), and NG-RC is (\ref{fig: mg_ngrc_welch}). The distributions for BEKK for Volterra kernel is (\ref{fig: bekk_volt_welch}), the polynomial kernel is (\ref{fig: bekk_poly_welch}), and NG-RC is (\ref{fig: bekk_ngrc_welch}). Only one dimension is displayed (dimension 118).}
		\label{fig: dist}
	\end{figure*}
	
	{We also observe that, especially in the case of BEKK, when significantly complex dynamical systems are being learned, considering large but finite lags or monomial degrees may be insufficient. Even if finite lags are sufficient, the Gram or feature matrix values may, anyway, be too large to be handled with finite precision. In such cases, the Volterra kernel regression significantly outperforms the other two methods because it is agnostic to the lags and monomial powers, and so its Gram values do not grow with the feature choice. Moreover, as we saw in Section~\ref{Infinite dimensional NG-RC and Volterra kernels}, taking infinite lag and monomial powers into consideration, offers a rich feature space and makes the associated RKHS universal, meaning that it can approximate complex systems to any desired accuracy.} 
	
	Overall, we find that the numerical simulations illustrate the points made in the theory discussed above. First, since the optimization problem in the NG-RC methodology is equivalent to that solved in polynomial kernel regression (as per Proposition~\ref{thm: ngrcispoly}), in more complex systems, it is better to kernelize to access computationally a richer feature space. This is illustrated especially by the superior performance of the polynomial kernel regression against the NG-RC in the Mackey-Glass system. Second, using the universal Volterra kernel, one can consider infinite lag and monomial powers, which is especially advantageous when learning significantly more complex systems, such as the BEKK input/output systems. In such cases, limited to finite lags and monomial orders, which need to be chosen (the Volterra kernel is agnostic to them), the polynomial kernel regression and NG-RC methodologies fail.
	
	\bigskip
	
	\noindent {\bf Acknowledgments:} The authors thank Daniel Gauthier for insightful discussions about the relation between NG-RC and the results in this paper. LG and JPO thank the hospitality of the Nanyang Technological University and the University of St.~Gallen, respectively; it is during respective visits to these two institutions that some of the results in this paper were obtained. HLJT is funded by a Nanyang President's Graduate Scholarship of Nanyang Technological University. JPO acknowledges partial financial support from the School of Physical and Mathematical Sciences of the Nanyang Technological University.
	
	\bibliographystyle{apsrev4-2_modified}
	\bibliography{GOLibrary}
	
\end{document}